\documentclass{article} % For LaTeX2e
\usepackage{iclr2027_conference,times}

\usepackage{amsmath,amsfonts,bm}

\def\eqref#1{equation~\ref{#1}}
\def\1{\bm{1}}

\DeclareMathAlphabet{\mathsfit}{\encodingdefault}{\sfdefault}{m}{sl}
\SetMathAlphabet{\mathsfit}{bold}{\encodingdefault}{\sfdefault}{bx}{n}

\PassOptionsToPackage{hyphens}{url}
\usepackage{hyperref}
\usepackage{url}
\usepackage[T1]{fontenc}
\usepackage{graphicx}
\usepackage{amsmath,amssymb}

\title{Persona Perspectives: Adopting Viewpoints \\  Reorganizes Representations in LLMs}

\author{Austin C.~Kozlowski, Avital Mintz \& Madison Lee %\thanks{ Use footnote for providing further information
\\
Knowledge Lab\\
University of Chicago\\
}

\iclrfinalcopy % Uncomment for camera-ready version, but NOT for submission.
\begin{document}

\maketitle

\begin{abstract}
Common methods for monitoring features in large language models (LLMs) such as linear probes and sparse autoencoders (SAEs) assume that features maintain a consistent representation across contexts. LLMs, however, can flexibly adopt different perspectives or ``personas'' that suggest different underlying systems of conceptual relations. In this paper, we examine whether prompting an LLM to adopt a persona alters the representational geometry of the model's latent space. We extract representations of over 9,000 common words and 33 semantic axes from models adopting liberal and conservative political personas and compare these against two neutral baselines. We find that persona conditioning systematically alters both the projections of words onto semantic axes and the alignments among the axes themselves in directions consistent with the specified ideology. We then test the implications of this reorganization for two safety-relevant features: truthfulness and morality. Linear probes, SAE-based classifiers, and classifiers built from manually selected word representations, all trained without a persona, produce divergent results when applied to models conditioned on political personas, producing politically slanted ratings of truth and morality on controversial statements. These findings suggest that internal representations, as they are conventionally extracted, can exhibit different geometric relations when the model adopts personas, potentially undermining monitoring efforts.
\end{abstract}

\section{Introduction}

The representations in large language models' (LLM) latent spaces are organized according to their semantic and pragmatic relations. Movements along linear features often correspond to consistent changes in meaning, and the geometric alignments between linear features reflect semantic similarity \citep{park2023linear, kozlowski2026semantic}. Thus, low-dimensional projections of latent space can at times serve as a map of an LLM's conceptual space, rendering visible the systems of similarities and differences the model utilizes in the service of token generation \citep{templeton2024scaling, li2025geometry}.

Static word embeddings like those generated by word2vec or GloVe have long been used to derive renderings of conceptual relations from text \citep{mikolov2013distributed, pennington2014glove, garg2018word, caliskan2017semantics}. Moreover, given their relatively modest training data requirements, static word embeddings can be trained on distinct corpora representative of specific discourses, and differences in representational geometries can be compared for cross-cultural analysis \citep{kozlowski2019geometry, nelson2021leveraging, stoltz2021cultural}. LLMs, by contrast, are rarely trained on texts representing a single perspective. Models are instead trained on large corpora comprising a multitude of perspectives, and these perspectives or ``personas'' are contextually evoked from the model through prompting \citep{argyle2023out, anthis2025position, bisbee2024synthetic, park2024generative}.

If cultural worldviews are reproduced through different representational geometries, and LLMs can invoke different perspectives through prompting, this raises the question of whether prompting reorganizes the LLM's representation space in accordance with the perspective being simulated. Previous work on the relation of personas to latent space has largely focused on representing personas as single vectors to be used for steering \citep{chen2025persona} or to be classified along a latent ``assistant axis'' \citep{lu2026assistant}. Little work has explored whether personas modify downstream relations between other features to instantiate the appropriate conceptual system.

This question has practical implications for efforts to monitor models' internal states. Monitoring is commonly accomplished with learned linear probes or sparse autoencoders which identify directions in latent space that correspond to concepts such as truthfulness, deception, or harmfulness\citep{marks2024geometry, goldowskydill2025detecting, bricken2023monosemanticity}. This approach rests on the assumptions that both the monitored concept and the object being classified keep consistent representations across contexts. If personas reorganize representational geometry, then a monitor extracted under one perspective may not transfer to another. We suggest that a statement represented as false or harmful under the assistant persona could be represented as true or beneficial by a model adopting a more sympathetic perspective, thus bypassing the monitor.

\section{Methods}

\paragraph{Models and personas.} To assess the effect of personas on representational geometry, we prompt LLMs to adopt politically liberal and conservative perspectives. We study political personas because they correspond to one of the greatest cultural cleavages in the United States today, and prior research suggests that they should manifest different representational geometries when encoded in language \citep{rim2026natural}. Moreover, liberals and conservatives differ meaningfully in their judgments of truth and morality, two domains with important implications for AI safety and alignment \citep{graham2009liberals, haidt2012righteous, lakoff2002moral}.

We induce liberal and conservative personas with simple system prompts: ``Respond as an American liberal Democrat'' or ``Respond as an American conservative Republican.'' We compare these against two baselines: an empty system prompt and a prompt instructing the model to ``Respond as a helpful AI assistant.'' Main analyses use the instruction-tuned Llama 3.1 8B \citep{grattafiori2024llama}. We replicate the analyses on seven further instruction-tuned models from four families, ranging from 4B to 70B parameters: Llama 3.1 70B, Gemma 3 4B and 27B \citep{Kamath2025gemma}, Qwen 3 8B and 32B \citep{yang2025qwen3}, and Ministral 3 8B and 14B \citep{liu2026ministral} (Appendix~\ref{app:models}).

\begin{figure}[t]
\centering
\includegraphics[width=\textwidth]{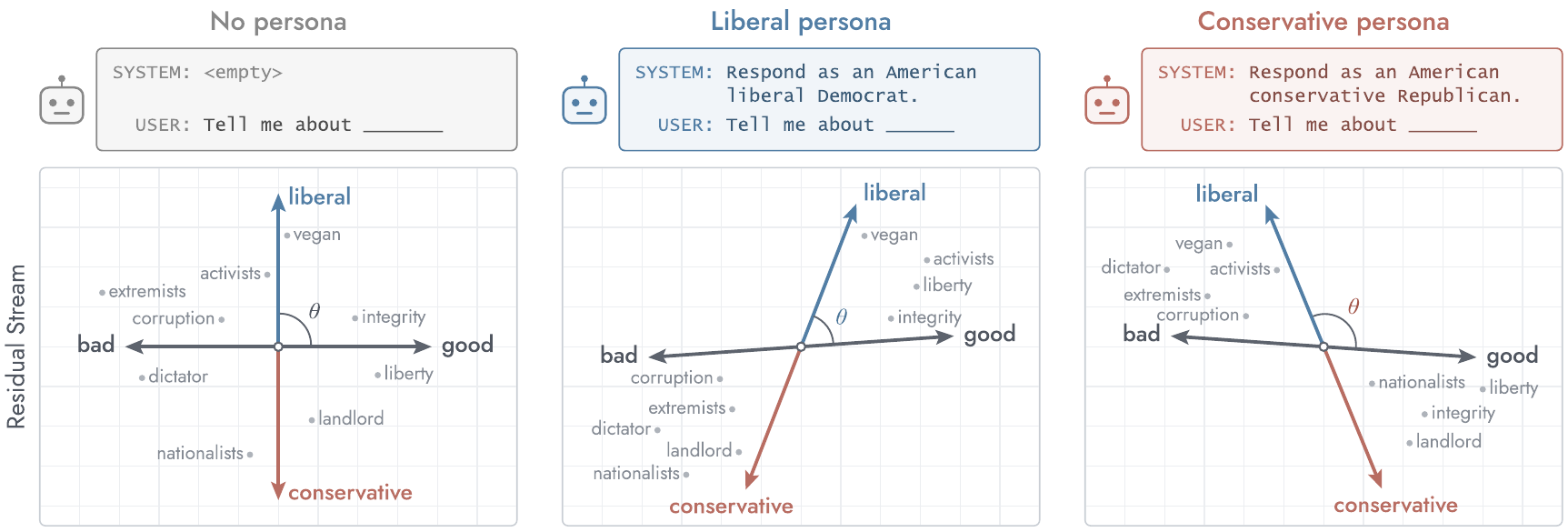}
\caption{Schematic diagram of three persona conditions and corresponding representation spaces. We hypothesize that persona conditions rotate semantic axes and reposition word representations in accordance with the persona's conceptual and linguistic relations.}
\label{fig:1}
\end{figure}

\paragraph{Word representations.} We build a ``feature bank'' of word representations from the 10,000 most common English words according to the wordfreq library \citep{speer2022wordfreq}. Each word \(w\) is inserted into five prompt templates, four posed as user messages and one with a pre-filled assistant response:

\begin{quote}
\textbf{USER:} When you hear ``\{\textit{word\(_i\)}\}'', what comes to mind?

\textbf{USER:} What does ``\{\textit{word\(_i\)}\}'' mean to you?

\textbf{USER:} Tell me your associations with ``\{\textit{word\(_i\)}\}.''

\textbf{USER:} Tell me about the term ``\{\textit{word\(_i\)}\}.''

\textbf{USER:} Define ``\{\textit{word\(_i\)}\}''. \textbf{ASSISTANT:} The term ``\{\textit{word\(_i\)}\}'' means
\end{quote}

The templates are based on \citet{chen2025persona}, and \citet{kozlowski2026semantic} showed that word representations extracted this way recover human-like semantic associations. Each prompt is run under each of the four system-prompt conditions \(c\). From each forward pass we record the residual-stream activation at the final token of the prompt, and we average these activations across the five templates to obtain a single representation of the word for each condition and layer \(\ell\).\footnote{For Llama 3.1 8B at layer 24, the liberal-minus-conservative differences in cosine similarity between semantic axes (all 528 axis pairs) constructed from last-token activations correlate at r = 0.95 with those constructed from activations mean-pooled over all non-system-prompt tokens, although the differences under mean pooling were on average roughly 50\% smaller (mean absolute difference 0.048 versus 0.099).} Activations are saved at every fourth layer. A sweep across depths showed persona-consistent representations form after 50\% depth (Appendix~\ref{app:depth}). We therefore use layer 24 (75\% depth) as the focus of our analyses, although analyses of nearby layers produce consistent results. After removing stop words and the anchor terms used to build the semantic axes, 9,493 words remain for analysis.

\paragraph{Semantic axes.} We construct 33 semantic axes following the approach of \citet{kozlowski2026semantic}, which finds high correlation between the geometric alignments of these semantic axes and human ratings. This method applies a contrastive-pair design in which axes are constructed by averaging the differences between representations of "anchor" terms designating each pole of the axis \citep{panickssery2023steering, tigges2023linear}. Each anchor term is represented by the same procedure used for the words above, and each axis \(f\) is the mean of the differences between the positive and negative anchor representations of its ten pairs:

\[
\mathbf{a}^{(\ell)}_{f,c} = \frac{1}{10} \sum_{j=1}^{10} \left( \mathbf{h}^{(\ell)}_{c}(\text{pos}_{f,j}) - \mathbf{h}^{(\ell)}_{c}(\text{neg}_{f,j}) \right).
\]

Axes are constructed separately within each system-prompt condition, so each persona has its own set of 33 axes. From these representations we compute two kinds of relations within each condition. The \emph{geometric alignment} of two axes \(f\) and \(g\) is their cosine similarity, \(\cos(\mathbf{a}_{f,c}, \mathbf{a}_{g,c})\), which captures how the two concepts are related to one another. The \emph{normalized projection} of a word onto an axis is \(\cos(\mathbf{h}_{c}(w), \mathbf{a}_{f,c})\), which captures where the word falls along that dimension.

\paragraph{Comparison with political word embeddings.} To test whether the persona-induced changes correspond specifically to liberal and conservative conceptual structure, we compare them with static word embeddings trained on politically slanted texts. We train separate word2vec models \citep{mikolov2013distributed, rehurek2010software} on liberal and conservative corpora comprised of partisan news, podcasts \citep{rizwan2025toxicity,wirtschafter2023audible}, and campaign emails \citep{willis2026political}. Details about the training data and model hyperparameters are provided in Appendix~\ref{app:word2vec}.

We construct the same 33 semantic axes in the word2vec models. However, because static embeddings rely on exact token matches, each anchor term needs many
occurrences in the training corpus to be estimated reliably, we use a larger set of anchor terms for the static embedding's axes. We draw these terms from \citet{boutyline2026forging} and filter to those that occur at least five times in both corpora, resulting in a median of 34 pairs per axis.

Because the two personas' representation spaces remain highly similar overall (for example, ``lawyer'' stays close to ``attorney'' and \textit{kind--cruel} stays close to \textit{good--bad} under either persona), we focus on their \emph{differences} rather than their raw geometry. For each system, we compute the 33 $\times$ 33 matrix of axis cosine similarities under the liberal and under the conservative condition and subtract the conservative matrix from the liberal matrix. For word2vec, the liberal and conservative models play the role of the two personas, and we average the difference matrices across three random training seeds. We then correlate two systems' difference matrices across the 528 unique axis pairs. We also apply this comparison between pairs of LLMs to test whether different models reorganize their geometry in the same way when adopting political personas.

\paragraph{Steering and spillovers.} To test whether the reorganized geometry has behavioral consequences, we adapt the steering design of \citet{kozlowski2026semantic}. By this design, we steer model activations along a semantic axis \(f\), but then test for behavioral effects along \textit{another} axis \(g\). For example, one set of tests would steer along \textit{soft--hard} while the model is rating a word as "good" or "bad." To induce the model to make a behavioral assessment of semantic associations, we use the following prompt template:

\begin{quote}
\textbf{USER:} Do you associate the term ``\{\textit{word\(_i\)}\}'' more with \{\textit{pos\(_g\)}\} or \{\textit{neg\(_g\)}\}? Respond with exactly one of those two words.
\end{quote}

 We then collect the log probabilities for the next generated token, summing over lowercase and capitalized forms of the given semantic axis's anchor terms (e.g. "good" and "bad"). We calculate the difference between the log probabilities for each anchor, resulting in log odds: \(L_g(w) = \log P(\text{pos}_g) - \log P(\text{neg}_g)\).

Using a forward hook on the residual stream after layer 24, we modify the hidden state \(\mathbf{x}\) at the final prompt token as:

\[
\mathbf{x}' = \mathbf{x} \pm \alpha \lVert \mathbf{x} \rVert \, \hat{\mathbf{a}}_{f},
\]

where \(\hat{\mathbf{a}}_{f}\) is the unit-normalized axis vector and \(\alpha = 0.33\) sets the intervention's magnitude as a fraction of the local residual-stream norm, giving every axis a proportionally equal intervention regardless of the raw magnitude of its vector. We average off-target effects for each pair of axes over 1,000 target words sampled at random from the common-word list. We intervene at the final prompt token, the position where persona effects are largest. Because this position directly produces the model's answer, we analyze only off-target pairs and exclude each axis's effect on its own rating. This leaves 1,056 directed off-target axis pairs. We run this full set of tests for each political persona and with the helpful AI assistant persona.

In models without personas, the alignment between two axes strongly predicts the spillover from one to the other \citep{kozlowski2026semantic}. We therefore test (1) whether each persona's spillovers are better predicted by its own axis alignments than by those of the other persona or the helpful assistant, and (2) whether differences in alignment between the personas predict differences in their spillovers. For example, if ``feminine'' aligns with ``good'' under the liberal persona but with ``bad'' under the conservative persona, then steering toward ``feminine'' should push ratings toward ``good'' under the liberal persona and toward ``bad'' under the conservative persona. If off-target effects are better predicted by the modified representational geometry, it suggests that the way one feature activation affects another's behavioral manifestation corresponds to their persona-conditioned geometric alignment.

\paragraph{Monitoring truthfulness and morality.} Finally, we examine how persona-conditioned geometry affects common monitoring techniques in two safety- and alignment-relevant domains. We compile two novel datasets of factually true/false statements and morally acceptable/unacceptable actions. The truthfulness dataset is constructed from high quality English Wikipedia articles (those designated "vital" by the editing community), spanning geography and places, science and technology, arts and culture, and biography. We used an LLM screen to remove articles on politically or socially controversial subjects. Using the content of each article, an LLM (GPT-5-mini) wrote four widely known, uncontroversial true claims and, for each, a minimally edited false counterpart (e.g., ``2001: A Space Odyssey is a 1968 science fiction film directed by Stanley Kubrick'' versus ``\ldots{} a 1968 romantic comedy directed by Stanley Kubrick''). The final dataset contains 9,192 statements in 4,596 true/false pairs from 1,149 articles; 230 complete articles are held out for testing (1,840 statements). 

The morality dataset contains 8,000 first-person actions in 4,000 matched acceptable/unacceptable pairs spanning 20 topics and 200 subtopics (e.g., ``I told my supervisor that I was genuinely sick and could not work my shift'' versus ``I lied to my supervisor that I was sick to avoid working my shift.''); complete pairs are held out for testing (400 actions from 200 pairs). These claims were initially generated using GPT-5-mini, then revised and cross-reviewed using Codex agents to improve clarity.

We train classifiers exclusively on these noncontroversial (un)truthful and (im)moral statements and only with an empty system prompt, and all probes read activations at the final prompt token before the model responds. We test three kinds of classifiers: (1) \emph{Learned linear probes} are logistic regressions on the full 4,096-dimensional residual stream. (2) \emph{SAE-feature probes} are logistic regressions on eight features selected from a pretrained sparse autoencoder at layer 19 \citep{arditi2025saes, bussmann2024batchtopk}. A concern with learned classifiers is that they may capture a correlated feature instead of the intended one, such as lexical differences between true and false statements \citep{levinstein2024still, farquhar2023challenges}. We therefore also test (3) \emph{word-representation probes} built from manually selected word vectors from the feature bank.\footnote{For truth: \emph{true, correct, accurate, factual, valid, verified, reliable} versus \emph{false, untrue, incorrect, inaccurate, erroneous, mistaken} and for morality: \emph{moral, ethical, virtuous, good, principled, honorable, just, fair} versus \emph{immoral, unethical, wicked, evil, dishonorable, unjust, unfair, inhumane}.} We learn weights for these anchors with a logistic classifier with coefficients constrained to be nonnegative.\footnote{Unlike the geometric analyses, probes are not fixed at 75\% depth. For Llama 3.1 8B, the layer of each linear and word-representation probe is selected by accuracy on held-out validation statements with no system prompt, and the SAE-feature probe uses the layer of the pretrained SAE (layer 19). For the other models in Appendix~\ref{app:probes}, a single probe layer is selected per model based on ordinary-statement accuracy. Probes at nearby layers produce similar results (Figure~\ref{fig:a_probe_depth}).}

We then apply these classifiers, trained on uncontroversial statements with no persona, to controversial statements with the political personas. The controversial truth set contains 360 statements on 12 topics, such as election integrity, climate change, and the COVID-19 pandemic; the morality set contains 300 first-person actions on 30 topics, such as abortion, gun ownership, and labor unions. Statements come in opposing pairs, one member of which we expect liberals to endorse and the other conservatives. These expectations are hypotheses about partisan endorsement (generated by GPT-5-mini), not ground truth. We hypothesize that a statement like ``massive voter fraud occurred in the 2020 election'' will be represented in the ``true'' direction when the model is prompted as a conservative but in the ``untrue'' direction when prompted as a liberal, so that a fixed truth probe will register the same claim as true under one persona and untrue under the other.

\section{Results}

\begin{figure}[t]
\centering
\includegraphics[width=\textwidth]{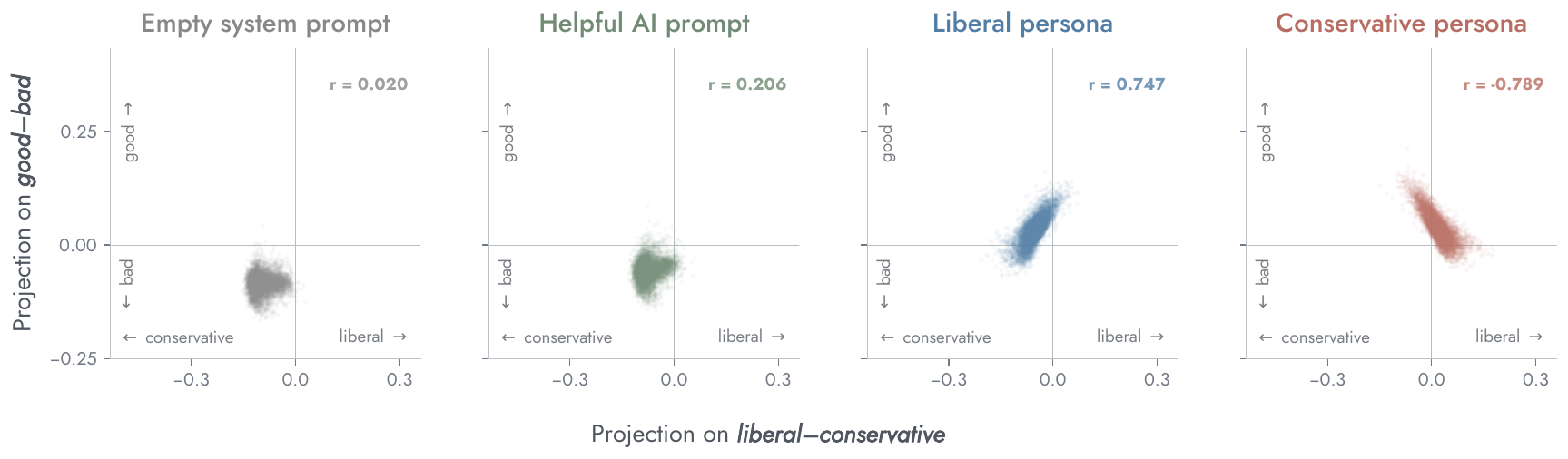}
\caption{Projections of 9,493 common words onto the liberal--conservative axis (horizontal) and the good--bad axis (vertical) under four system-prompt conditions, Llama 3.1 8B, layer 24. Each point is a word's representation vector. Both the word vectors and the axes are extracted within each condition. Pearson correlations between the two projections are shown in each panel.}
\label{fig:2}
\end{figure}

\paragraph{Personas reorganize semantic relations.} Figure~\ref{fig:2} displays the projections of word representations onto the \textit{liberal--conservative} and  \textit{good--bad} axes under all four persona conditions. Without a system prompt, the two projections are unrelated (r = .02), and the helpful-assistant prompt produces only a weak association (r = .21). Under the liberal persona, words that fall toward ``liberal'' also fall toward ``good'' (r = .75); under the conservative persona, the relation reverses, so that words falling toward ``conservative'' fall toward ``good'' (r = $-$.79).

\begin{figure}[t]
\centering
\includegraphics[width=\textwidth]{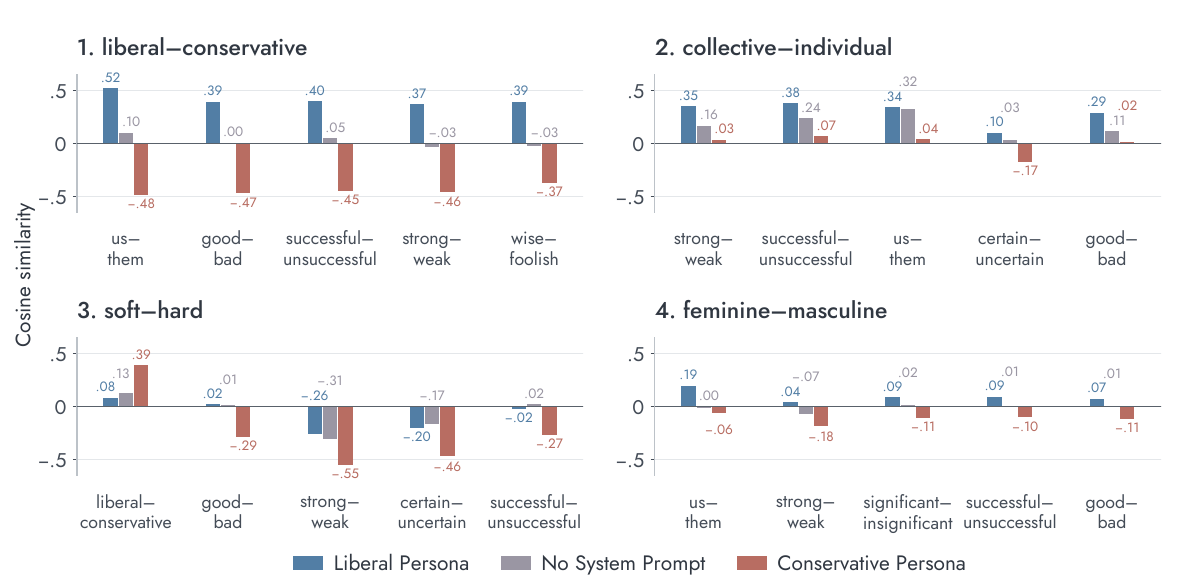}
\caption{The four axes whose relations to the other 32 axes change most between the liberal and conservative personas (ranked by mean absolute change in cosine similarity), each shown with the five partner axes exhibiting the greatest changes. Bars show the cosine similarity between the focal axis and each partner under different persona conditions. Llama 3.1 8B, layer 24.}
\label{fig:3}
\end{figure}

The same reorganization appears in the cosine similarities between the axes themselves. Figure~\ref{fig:3} shows the four axes whose relations to the other 32 axes change most between the liberal and conservative personas, each with their five partner axes that exhibit the greatest change. The \textit{liberal--conservative} axis changes by far the most. Under the liberal persona, \textit{liberal} aligns with \textit{us} (cosine .52), \textit{good} (.39), \textit{successful} (.40), \textit{strong} (.37), and \textit{wise} (.39). Under the conservative persona these relations reverse, so that \textit{conservative} aligns with the same poles ($-$.48, $-$.47, $-$.45, $-$.46, $-$.37). Without a system prompt, all five cosines are near zero ($-$.03 to .10). Partisan affect toward the parties themselves is the most easily anticipated difference between liberal and conservative judgment, and it is the strongest difference we identify.

The next largest changes involve \textit{collective--individual}, \textit{soft--hard}, and \textit{feminine--masculine}. Under the liberal persona, \textit{collective} aligns with \textit{strong} (.35), \textit{successful} (.38), \textit{us} (.34), and \textit{good} (.29); under the conservative persona, \textit{collective--individual} is nearly orthogonal to all of these (.02 to .07). The conservative persona reorganizes the \textit{soft--hard} axis: \textit{hard} aligns with \textit{conservative} (.39, versus .08 under the liberal persona), with \textit{good} ($-$.29 versus .02), and more strongly with \textit{strong} ($-$.55 versus $-$.26) and \textit{certain} ($-$.46 versus $-$.20). The liberal persona weakly aligns \textit{feminine} with \textit{us} (.19) and positive evaluative poles, whereas the conservative persona aligns \textit{masculine} with \textit{strong} ($-$.18), \textit{significant} ($-$.11), \textit{successful} ($-$.10), and \textit{good} ($-$.11). These differences corroborate a great deal of prior research in political psychology and public opinion that identifies  the greatest difference between ideological groups as their affect toward the partisan identities themselves \citep{iyengar2012affect, iyengar2019origins}, and other studies suggesting the liberal/conservative divide is grounded in egalitarianism versus individualism \citep{feldman1992political} as well as gender, not only explicitly but implicitly through emphases on strength and strictness versus gentleness and care \citep{box-steffensmeier2004dynamics, winter2010masculine, lakoff2002moral}.

\paragraph{Reorganization is shared across models and matches partisan text.} Beyond conforming to prior theories from the social sciences, we empirically test whether these changes reflect liberal and conservative conceptual structure in two ways: whether different LLMs reorganize their geometry in the same way, and whether the reorganization matches the differences between word embeddings trained on liberal and conservative texts. Both comparisons correlate liberal-minus-conservative differences in axis cosine similarities across the 528 unique axis pairs (see Methods).

\begin{figure}[t]
\centering
\includegraphics[width=\textwidth]{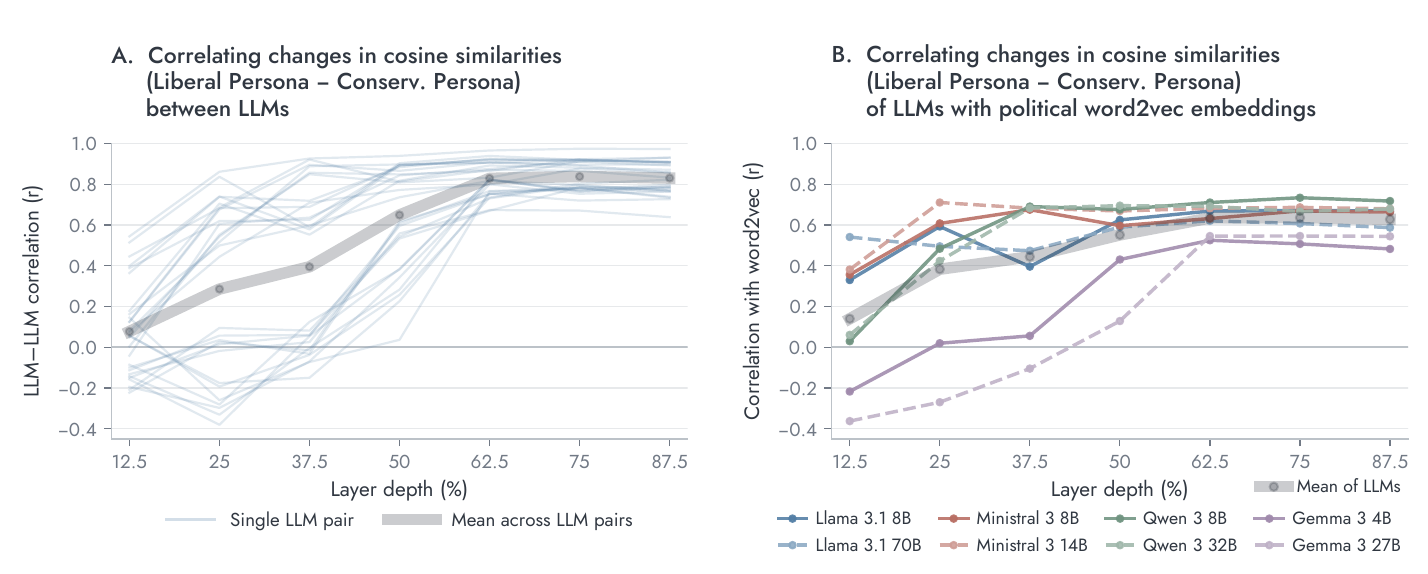}
\caption{Agreement in persona-induced geometry across layer depth. (A) Correlations between differences in cosine similarities of semantic axes between liberal and conservative personas for the 28 pairs of the 8 LLMs. (B) Correlation between each LLM's liberal-minus-conservative differences and the differences between word2vec models trained on liberal versus conservative texts}
\label{fig:4}
\end{figure}

Figure~\ref{fig:4}A shows that the differences between the liberal and conservative personas are highly similar across different models. In the later layers, the persona differences between the 8 tested models correlate strongly (mean r = .84 across the 28 model pairs at 75\% depth). It is important to note these are \textit{differences} between cosine similarities that are being correlated. Therefore, we are not capturing baseline representational similarity -- only differences that similarly emerge between liberal and conservative persona-conditioned representations. Figure~\ref{fig:4}B compares LLM persona differences with the differences between the liberal and conservative word2vec models. In the later layers, the correlation is consistently positive and substantial (mean r = .64 across the eight models at 75\% depth; r = .67 for Llama 3.1 8B at layer 24). These findings provisionally suggest that, when a persona is specified in an LLM's system prompt, it ``spins up'' a representation space resembling the one that would be obtained by training specifically on that perspective's texts.

\begin{figure}[t]
\centering
\includegraphics[width=\textwidth]{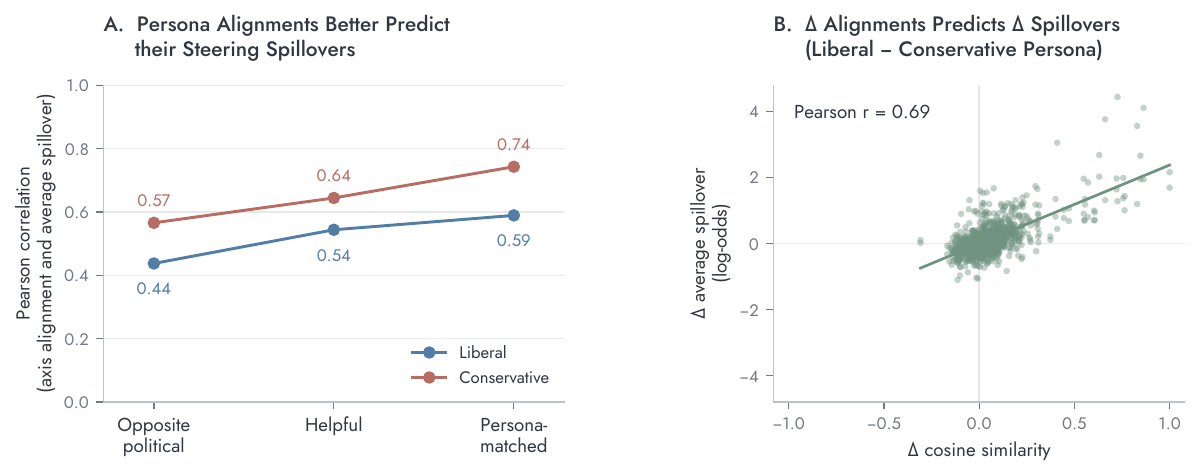}
\caption{Persona geometry predicts steering spillovers. (A) The spillover effects observed under a political persona condition are best predicted by the semantic axis alignments corresponding to that persona. (B) Difference in cosine similarities predicts difference in spillover effects, subtracting values of conservative persona from values of liberal persona. Each point is a directed steering-axis $\rightarrow$ rating-axis pair. Spillovers are averaged over 1,000 common words; Llama 3.1 8B, layer 24.}
\label{fig:5}
\end{figure}

\paragraph{Reorganized geometry predicts behavioral spillovers.} If the reorganized geometry reflects how each persona actually uses its concepts, it should predict the off-target effects of steering under that persona. Figure~\ref{fig:5}A shows evidence of this. The correlation between axis alignment and spillover is highest when a persona's spillovers are predicted from its own geometry (r = .59 for the liberal persona and .74 for the conservative persona), intermediate when predicted from the helpful-assistant geometry (.54 and .64), and lowest when predicted from the opposite persona's geometry (.44 and .57). Figure~\ref{fig:5}B shows that the \emph{differences} between the personas' geometries predict the differences in their spillovers (r = .69 across the 1,056 directed axis pairs): where the liberal persona aligns two axes more closely than the conservative persona does, steering along one axis produces a larger spillover onto the other under the liberal persona. Changes in representational geometry are thus accompanied by corresponding changes in how the model's judgments respond to intervention.

\begin{figure}[t]
\centering
\includegraphics[width=\textwidth]{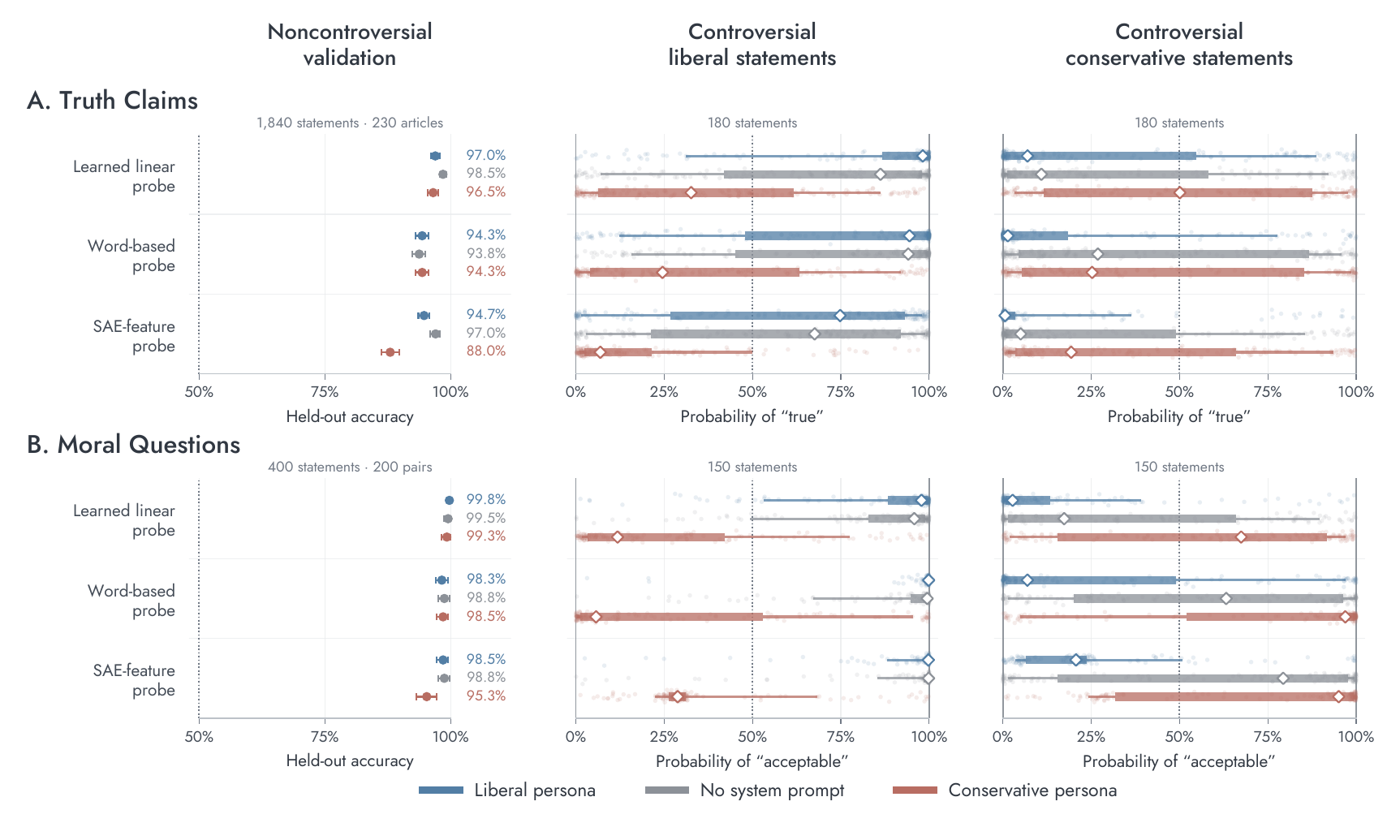}
\caption{Classifications of fixed truth and morality monitors under different personas. Left column: accuracy on held-out noncontroversial statements, 95\% bootstrap confidence intervals. Middle and right columns: probe-assigned probability that controversial statements conventionally endorsed by liberals (middle) or conservatives (right) are true or acceptable. Diamonds mark medians, thick bars the middle 50\% of statements, thin bars the middle 80\%, and faint points individual statements.}
\label{fig:6}
\end{figure}

\paragraph{Fixed truth and morality monitors give persona-dependent readings.} Finally, we examine the consequences of this reorganization for the three probes. Figure~\ref{fig:6} shows the results for truth claims (A) and moral questions (B). On held-out noncontroversial statements (left column), all three classifiers discriminate true from false statements (94--99\% accuracy with no system prompt) and acceptable from unacceptable actions (about 99\%). Accuracy on these ordinary items stays high under both personas, with one exception: the SAE-feature truth probe falls from 97\% to 88\% under the conservative persona. On conventional topics, then, the personas largely preserve the representations of truth and morality that the classifiers rely on.

The middle column shows controversial stances conventionally endorsed by liberals. With no system prompt, the probes classify most of these liberal stances as true or acceptable. This accords with behavior: in the absence of a system prompt, the model tends to endorse liberal positions and challenge conservative ones (Appendix~\ref{app:behavior}, Figure~\ref{fig:a_behavior_base}). The liberal persona raises these probabilities further for truth claims under the learned linear and SAE-feature probes (median 98\% versus 86\%, and 75\% versus 68\%). On moral questions, the probes already classify nearly all liberal stances as acceptable, leaving little room for the liberal persona to move them.

For the conservative persona, however, the median probability falls well below 50\% for all three probes for the controversial liberal truth claims. It is important to note that these are \emph{the same probes} in all three conditions: each points in the same absolute direction in the latent space. If controversial liberal stances are classified differently under the conservative persona, it is because the same statements activate a different region of the representation space: the truth claims shift toward the ``untrue'' side of the probe and the moral claims toward the ``unacceptable'' side.

We see similar effects for controversial conservative statements (right column). The probabilities of ``true'' are substantially lower under the liberal persona than under the conservative persona for all three probes. Many of these statements are classified as untrue even under the conservative persona, but this matches the model's behavior: despite the conservative persona, the model rarely takes some positions like denying the reality of climate change (Figure~\ref{fig:a_behavior_persona}). The effects are clearer still for controversial conservative moral stances: the median probability of ``acceptable'' is well below 50\% for all three probes under the liberal persona (3--21\%) and well above 50\% under the conservative persona (67--97\%).

\section{Related Literature}

\paragraph{LLM personas.} Personas have drawn attention from two main directions. Computational social scientists use persona prompts to simulate human respondents, asking whether models conditioned on demographic or ideological identities reproduce the opinions of the corresponding populations \citep{argyle2023out, santurkar2023whose, bisbee2024synthetic, park2024generative, anthis2025position}. This line of research has focused on the correspondence between model outputs and the outputs of humans they are intended to simulate, but has given less attention to the relation between personas and LLM internal representations. Alignment researchers, in turn, study the character that models adopt, motivated both by questions of what the default assistant persona should be \citep{marks2026persona} and by concerns that models can drift into misaligned personas \citep{li2024measuring, betley2025emergent, lu2026assistant}. This research typically treats the persona itself as a direction or subspace that can be read out or steered \citep{chen2025persona, wang2025persona, lu2026assistant}. We instead treat the persona as a context that reorganizes the relations among the model's other representations.

\paragraph{Linear features and monitoring.} Much of interpretability rests on the premise that concepts are represented as stable linear directions \citep{park2023linear, elhage2022superposition}, supported by evidence for linear representations of truth and other concepts \citep{marks2024geometry, zou2023representation}. This premise underwrites the probes and SAE features used to monitor models for deception, defection, and other safety-relevant states \citep{goldowskydill2025detecting, macdiarmid2024simple, bricken2023monosemanticity}. Yet growing evidence suggests that these representations are not fixed: truth probes shift with the premises and instructions in context \citep{schouten2025truthvalue, poulis2026testing} and degrade under distribution shift \citep{levinstein2024still, kramar2026building}, in-context structure can reorganize concept representations \citep{park2025iclr, lubana2025priors}, and factuality directions can flip over the course of a conversation as the model takes on a role \citep{lampinen2026linear}. We extend this line of work by showing that a single, benign instruction to adopt a perspective systematically reorganizes the geometry of many representations at once, in a direction that matches the corresponding ideology.

\section{Conclusion}

We present evidence that when models are instructed to adopt different perspectives in their system prompt, this systematically restructures the model's representation space to induce conceptual relations associated with that perspective. This restructuring is shared across models, mirrors the conceptual differences found in partisan text, and carries over both to the model's behavior and to the monitors built on its representations.

These findings provide an alternative to conceptualizing personas as mere directions added to an otherwise fixed representation space. We find that a one-sentence instruction reorganizes the relations among the model's representations to compose the conceptual relations characteristic of that perspective. This encourages analysts to locate the simulation of perspectives \citep{shanahan2023role} in the geometry of the latent space as well as in generated outputs.

Moreover, they motivate a reconsideration of methods that conceive of representations as context-invariant directions in latent space. A given concept, whether it is an attribute like "liberal" or an entity like "the Democratic Party," can operate differently for different perspectives. It might be activated by different associations and it might go on to activate different associations in turn. Although we do not dismiss the possibility of training a classifier that functions well across contexts, our work suggests that simple approaches to probe construction, which do not account for the multiplicity of perspectives an LLM learns, are susceptible to failure under changing personas.

\bibliography{iclr2027_conference}

@article{lubana2025priors,
  title         = {Priors in Time: Missing Inductive Biases for Language Model Interpretability},
  author        = {Lubana, Ekdeep Singh and Rager, Can and Hindupur, Sai Sumedh R. and Costa, Val{\'e}rie and Tuckute, Greta and Patel, Oam and Murthy, Sonia Krishna and Fel, Thomas and Wurgaft, Daniel and Bigelow, Eric J. and Lin, Johnny and Ba, Demba and Wattenberg, Martin and Vi{\'e}gas, Fernanda and Weber, Melanie and Mueller, Aaron},
  journal       = {arXiv preprint arXiv:2511.01836},
  year          = {2025},
  eprint        = {2511.01836},
  archivePrefix = {arXiv},
  primaryClass  = {cs.LG}
}

@article{li2025geometry,
  title={The geometry of concepts: Sparse autoencoder feature structure},
  author={Li, Yuxiao and Michaud, Eric J and Baek, David D and Engels, Joshua and Sun, Xiaoqing and Tegmark, Max},
  journal={Entropy},
  volume={27},
  number={4},
  pages={344},
  year={2025},
  publisher={MDPI}
}

@article{templeton2024scaling,
   title={Scaling Monosemanticity: Extracting Interpretable Features from Claude 3 Sonnet},
   author={Templeton, Adly and Conerly, Tom and Marcus, Jonathan and Lindsey, Jack and Bricken, Trenton and Chen, Brian and Pearce, Adam and Citro, Craig and Ameisen, Emmanuel and Jones, Andy and Cunningham, Hoagy and Turner, Nicholas L and McDougall, Callum and MacDiarmid, Monte and Freeman, C. Daniel and Sumers, Theodore R. and Rees, Edward and Batson, Joshua and Jermyn, Adam and Carter, Shan and Olah, Chris and Henighan, Tom},
   year={2024},
   journal={Transformer Circuits Thread},
   url={https://transformer-circuits.pub/2024/scaling-monosemanticity/index.html}
}

@inproceedings{park2025iclr,
  title     = {{ICLR}: In-Context Learning of Representations},
  author    = {Park, Core Francisco and Lee, Andrew and Lubana, Ekdeep Singh and Yang, Yongyi and Okawa, Maya and Nishi, Kento and Wattenberg, Martin and Tanaka, Hidenori},
  booktitle = {The Thirteenth International Conference on Learning Representations},
  year      = {2025}
}

@article{lampinen2026linear,
  title         = {Linear Representations in Language Models Can Change Dramatically over a Conversation},
  author        = {Lampinen, Andrew Kyle and Li, Yuxuan and Hosseini, Eghbal and Bhardwaj, Sangnie and Shanahan, Murray},
  journal       = {arXiv preprint arXiv:2601.20834},
  year          = {2026},
  eprint        = {2601.20834},
  archivePrefix = {arXiv},
  primaryClass  = {cs.CL}
}

@inproceedings{marks2024geometry,
  title     = {The Geometry of Truth: Emergent Linear Structure in Large Language Model Representations of True/False Datasets},
  author    = {Marks, Samuel and Tegmark, Max},
  booktitle = {First Conference on Language Modeling},
  year      = {2024},
  url       = {https://openreview.net/forum?id=aajyHYjjsk}
}

@inproceedings{goldowskydill2025detecting,
  title     = {Detecting Strategic Deception with Linear Probes},
  author    = {Goldowsky-Dill, Nicholas and Chughtai, Bilal and Heimersheim, Stefan and Hobbhahn, Marius},
  booktitle = {Proceedings of the 42nd International Conference on Machine Learning},
  series    = {Proceedings of Machine Learning Research},
  volume    = {267},
  pages     = {19755--19786},
  publisher = {PMLR},
  year      = {2025},
  url       = {https://proceedings.mlr.press/v267/goldowsky-dill25a.html}
}

@article{bricken2023monosemanticity,
  title   = {Towards Monosemanticity: Decomposing Language Models With Dictionary Learning},
  author  = {Bricken, Trenton and Templeton, Adly and Batson, Joshua and Chen, Brian and Jermyn, Adam and Conerly, Tom and Turner, Nicholas L. and Anil, Cem and Denison, Carson and Askell, Amanda and Lasenby, Robert and Wu, Yifan and Kravec, Shauna and Schiefer, Nicholas and Maxwell, Tim and Joseph, Nicholas and Tamkin, Alex and Nguyen, Karina and McLean, Brayden and Burke, Josiah E. and Hume, Tristan and Carter, Shan and Henighan, Tom and Olah, Christopher},
  journal = {Transformer Circuits Thread},
  year    = {2023},
  url     = {https://transformer-circuits.pub/2023/monosemantic-features}
}

@article{Kamath2025gemma,
  title={Gemma 3 technical report},
  author={Kamath, Aishwarya and Ferret, Johan and Pathak, Shreya and Vieillard, Nino and Merhej, Ramona and Perrin, Sarah and Matejovicova, Tatiana and Ram{\'e}, Alexandre and Rivi{\`e}re, Morgane and others},
  journal={arXiv preprint arXiv:2503.19786},
  year={2025}
}

@inproceedings{anthis2025position,
  author    = {Anthis, Jacy Reese and Liu, Ryan and Richardson, Sean M. and Kozlowski, Austin C. and Koch, Bernard and Brynjolfsson, Erik and Evans, James and Bernstein, Michael S.},
  title     = {Position: {LLM} Social Simulations Are a Promising Research Method},
  booktitle = {Proceedings of the 42nd International Conference on Machine Learning},
  series    = {Proceedings of Machine Learning Research},
  volume    = {267},
  pages     = {81005--81034},
  publisher = {PMLR},
  year      = {2025},
  editor    = {Singh, Aarti and Fazel, Maryam and Hsu, Daniel and Lacoste-Julien, Simon and Berkenkamp, Felix and Maharaj, Tegan and Wagstaff, Kiri and Zhu, Jerry},
  url       = {https://proceedings.mlr.press/v267/anthis25a.html}
}

@misc{arditi2025saes,
  author       = {Arditi, Andy},
  title        = {saes-llama-3.1-8b-instruct: {BatchTopK} Residual-Stream Sparse Autoencoders for {Llama}~3.1 8B Instruct},
  year         = {2025},
  howpublished = {Hugging Face model repository},
  url          = {https://huggingface.co/andyrdt/saes-llama-3.1-8b-instruct}
}

@article{argyle2023out,
  author  = {Argyle, Lisa P. and Busby, Ethan C. and Fulda, Nancy and Gubler, Joshua R. and Rytting, Christopher and Wingate, David},
  title   = {Out of One, Many: Using Language Models to Simulate Human Samples},
  journal = {Political Analysis},
  year    = {2023},
  volume  = {31},
  number  = {3},
  pages   = {337--351},
  doi     = {10.1017/pan.2023.2}
}

@article{bisbee2024synthetic,
  author  = {Bisbee, James and Clinton, Joshua D. and Dorff, Cassy and Kenkel, Brenton and Larson, Jennifer M.},
  title   = {Synthetic Replacements for Human Survey Data? {T}he Perils of Large Language Models},
  journal = {Political Analysis},
  year    = {2024},
  volume  = {32},
  number  = {4},
  pages   = {401--416},
  doi     = {10.1017/pan.2024.5}
}

@article{box-steffensmeier2004dynamics,
  author  = {Box-Steffensmeier, Janet M. and De Boef, Suzanna and Lin, Tse-Min},
  title   = {The Dynamics of the Partisan Gender Gap},
  journal = {American Political Science Review},
  volume  = {98},
  number  = {3},
  pages   = {515--528},
  year    = {2004},
  doi     = {10.1017/S0003055404001315}
}

@misc{bussmann2024batchtopk,
  author        = {Bussmann, Bart and Leask, Patrick and Nanda, Neel},
  title         = {{BatchTopK} Sparse Autoencoders},
  year          = {2024},
  eprint        = {2412.06410},
  archivePrefix = {arXiv},
  primaryClass  = {cs.LG},
  url           = {https://arxiv.org/abs/2412.06410}
}

@article{caliskan2017semantics,
  title={Semantics derived automatically from language corpora contain human-like biases},
  author={Caliskan, Aylin and Bryson, Joanna J and Narayanan, Arvind},
  journal={Science},
  volume={356},
  number={6334},
  pages={183--186},
  year={2017},
  publisher={American Association for the Advancement of Science}
}

@misc{chen2025persona,
  author        = {Chen, Runjin and Arditi, Andy and Sleight, Henry and Evans, Owain and Lindsey, Jack},
  title         = {Persona Vectors: Monitoring and Controlling Character Traits in Language Models},
  year          = {2025},
  eprint        = {2507.21509},
  archivePrefix = {arXiv},
  primaryClass  = {cs.CL},
  url           = {https://arxiv.org/abs/2507.21509}
}

@misc{farquhar2023challenges,
  author        = {Farquhar, Sebastian and Varma, Vikrant and Kenton, Zachary and Gasteiger, Johannes and Mikulik, Vladimir and Shah, Rohin},
  title         = {Challenges with Unsupervised {LLM} Knowledge Discovery},
  year          = {2023},
  eprint        = {2312.10029},
  archivePrefix = {arXiv},
  primaryClass  = {cs.LG},
  doi           = {10.48550/arXiv.2312.10029}
}

@article{garg2018word,
  title={Word embeddings quantify 100 years of gender and ethnic stereotypes},
  author={Garg, Nikhil and Schiebinger, Londa and Jurafsky, Dan and Zou, James},
  journal={Proceedings of the National Academy of Sciences},
  volume={115},
  number={16},
  pages={E3635--E3644},
  year={2018},
  publisher={National Academy of Sciences}
}

@article{graham2009liberals,
  author  = {Graham, Jesse and Haidt, Jonathan and Nosek, Brian A.},
  title   = {Liberals and Conservatives Rely on Different Sets of Moral Foundations},
  journal = {Journal of Personality and Social Psychology},
  volume  = {96},
  number  = {5},
  pages   = {1029--1046},
  year    = {2009},
  doi     = {10.1037/a0015141}
}

@article{grattafiori2024llama,
  title={The llama 3 herd of models},
  author={Grattafiori, Aaron and Dubey, Abhimanyu and Jauhri, Abhinav and Pandey, Abhinav and Kadian, Abhishek and Al-Dahle, Ahmad and Letman, Aiesha and Mathur, Akhil and Schelten, Alan and Vaughan, Alex and others},
  journal={arXiv preprint arXiv:2407.21783},
  year={2024}
}

@book{haidt2012righteous,
  author    = {Haidt, Jonathan},
  title     = {The Righteous Mind: Why Good People Are Divided by Politics and Religion},
  publisher = {Pantheon Books},
  address   = {New York},
  year      = {2012},
  isbn      = {9780307377906}
}

@article{iyengar2012affect,
  author  = {Iyengar, Shanto and Sood, Gaurav and Lelkes, Yphtach},
  title   = {Affect, Not Ideology: A Social Identity Perspective on Polarization},
  journal = {Public Opinion Quarterly},
  volume  = {76},
  number  = {3},
  pages   = {405--431},
  year    = {2012},
  doi     = {10.1093/poq/nfs038},
  note    = {Erratum: Public Opinion Quarterly 76(4):819, doi:10.1093/poq/nfs059}
}

@article{iyengar2019origins,
  author  = {Iyengar, Shanto and Lelkes, Yphtach and Levendusky, Matthew and Malhotra, Neil and Westwood, Sean J.},
  title   = {The Origins and Consequences of Affective Polarization in the {United States}},
  journal = {Annual Review of Political Science},
  volume  = {22},
  number  = {1},
  pages   = {129--146},
  year    = {2019},
  doi     = {10.1146/annurev-polisci-051117-073034}
}

@article{kozlowski2019geometry,
  title={The geometry of culture: Analyzing the meanings of class through word embeddings},
  author={Kozlowski, Austin C. and Taddy, Matt and Evans, James A},
  journal={American Sociological Review},
  volume={84},
  number={5},
  pages={905--949},
  year={2019},
  publisher={SAGE Publications Sage CA: Los Angeles, CA}
}

@inproceedings{kozlowski2026semantic,
  author        = {Kozlowski, Austin C. and Boutyline, Andrei},
  title         = {Semantic Structure of Feature Space in Large Language Models},
  booktitle     = {Conference on Language Modeling (COLM)},
  year          = {2026},
  eprint        = {2604.27169},
  archivePrefix = {arXiv},
  url           = {https://arxiv.org/abs/2604.27169}
}

@book{lakoff2002moral,
  author    = {Lakoff, George},
  title     = {Moral Politics: How Liberals and Conservatives Think},
  edition   = {2},
  publisher = {University of Chicago Press},
  address   = {Chicago},
  year      = {2002},
  doi       = {10.7208/chicago/9780226471006.001.0001}
}

@article{levinstein2024still,
  author  = {Levinstein, Benjamin A. and Herrmann, Daniel A.},
  title   = {Still No Lie Detector for Language Models: Probing Empirical and Conceptual Roadblocks},
  journal = {Philosophical Studies},
  volume  = {182},
  number  = {7},
  pages   = {1539--1565},
  year    = {2024},
  doi     = {10.1007/s11098-023-02094-3},
  note    = {Published online 17 February 2024; issue dated 2025}
}

@misc{liu2026ministral,
  author        = {Liu, Alexander H. and Khandelwal, Kartik and Subramanian, Sandeep and others},
  title         = {Ministral 3},
  year          = {2026},
  eprint        = {2601.08584},
  archivePrefix = {arXiv},
  primaryClass  = {cs.CL},
  url           = {https://arxiv.org/abs/2601.08584}
}

@misc{lu2026assistant,
  author        = {Lu, Christina and Gallagher, Jack and Michala, Jonathan and Fish, Kyle and Lindsey, Jack},
  title         = {The Assistant Axis: Situating and Stabilizing the Default Persona of Language Models},
  year          = {2026},
  eprint        = {2601.10387},
  archivePrefix = {arXiv},
  primaryClass  = {cs.CL},
  url           = {https://arxiv.org/abs/2601.10387}
}

@article{mikolov2013distributed,
  title={Distributed representations of words and phrases and their compositionality},
  author={Mikolov, Tomas and Sutskever, Ilya and Chen, Kai and Corrado, Greg S and Dean, Jeff},
  journal={Advances in neural information processing systems},
  volume={26},
  year={2013}
}

@article{nelson2021leveraging,
  author  = {Nelson, Laura K.},
  title   = {Leveraging the Alignment between Machine Learning and Intersectionality: Using Word Embeddings to Measure Intersectional Experiences of the Nineteenth Century {U.S.} {South}},
  journal = {Poetics},
  volume  = {88},
  pages   = {101539},
  year    = {2021},
  doi     = {10.1016/j.poetic.2021.101539}
}

@article{park2023linear,
  title={The linear representation hypothesis and the geometry of large language models},
  author={Park, Kiho and Choe, Yo Joong and Veitch, Victor},
  journal={arXiv preprint arXiv:2311.03658},
  year={2023}
}

@misc{park2024generative,
  author        = {Park, Joon Sung and Zou, Carolyn Q. and Shaw, Aaron and Hill, Benjamin Mako and Cai, Carrie and Morris, Meredith Ringel and Willer, Robb and Liang, Percy and Bernstein, Michael S.},
  title         = {Generative Agent Simulations of 1,000 People},
  year          = {2024},
  eprint        = {2411.10109},
  archivePrefix = {arXiv},
  primaryClass  = {cs.AI},
  url           = {https://arxiv.org/abs/2411.10109}
}

@inproceedings{pennington2014glove,
  title={Glove: Global vectors for word representation},
  author={Pennington, Jeffrey and Socher, Richard and Manning, Christopher D},
  booktitle={Proceedings of the 2014 conference on empirical methods in natural language processing (EMNLP)},
  pages={1532--1543},
  year={2014}
}

@inproceedings{rehurek2010software,
  author    = {{\v{R}}eh{\r{u}}{\v{r}}ek, Radim and Sojka, Petr},
  title     = {Software Framework for Topic Modelling with Large Corpora},
  booktitle = {Proceedings of the {LREC} 2010 Workshop on New Challenges for {NLP} Frameworks},
  pages     = {45--50},
  publisher = {ELRA},
  address   = {Valletta, Malta},
  year      = {2010},
  month     = may
}

@article{rim2026natural,
  author  = {Rim, Nakwon and Jackson, Joshua Conrad and Berman, Marc G. and Leong, Yuan Chang},
  title   = {Natural Language Reveals that Political Partisans Are More Affectively Aligned over Political Issues than Partisan Identities},
  journal = {Communications Psychology},
  year    = {2026},
  volume  = {4},
  number  = {1},
  pages   = {65},
  doi     = {10.1038/s44271-026-00430-x},
  url     = {https://www.nature.com/articles/s44271-026-00430-x}
}

@inproceedings{rizwan2025toxicity,
  author    = {Rizwan, Naquee and Deb, Nayandeep and Roy, Sarthak and Solanki, Vishwajeet Singh and Garimella, Kiran and Mukherjee, Animesh},
  title     = {Toxicity Begets Toxicity: Unraveling Conversational Chains in Political Podcasts},
  booktitle = {Proceedings of the 33rd ACM International Conference on Multimedia},
  series    = {MM '25},
  publisher = {Association for Computing Machinery},
  address   = {New York, NY, USA},
  pages     = {11776--11784},
  year      = {2025},
  doi       = {10.1145/3746027.3754553}
}

@article{stoltz2021cultural,
  author  = {Stoltz, Dustin S. and Taylor, Marshall A.},
  title   = {Cultural Cartography with Word Embeddings},
  journal = {Poetics},
  volume  = {88},
  pages   = {101567},
  year    = {2021},
  doi     = {10.1016/j.poetic.2021.101567}
}

@misc{willis2026political,
  author       = {Willis, Derek},
  title        = {Political Email Archive},
  year         = {2026},
  howpublished = {\url{https://thescoop.org/political-emails/}},
  note         = {Archive of U.S. political fundraising emails, 2015--present. Code and data: \url{https://github.com/dwillis/political-emails}. Accessed September 2026}
}

@article{winter2010masculine,
  author  = {Winter, Nicholas J. G.},
  title   = {Masculine {Republicans} and Feminine {Democrats}: Gender and {Americans'} Explicit and Implicit Images of the Political Parties},
  journal = {Political Behavior},
  volume  = {32},
  number  = {4},
  pages   = {587--618},
  year    = {2010},
  doi     = {10.1007/s11109-010-9131-z}
}

@techreport{wirtschafter2023audible,
  author      = {Wirtschafter, Valerie},
  title       = {Audible Reckoning: How Top Political Podcasters Spread Unsubstantiated and False Claims},
  institution = {Brookings Institution},
  address     = {Washington, DC},
  year        = {2023},
  month       = feb,
  url         = {https://www.brookings.edu/articles/audible-reckoning-how-top-political-podcasters-spread-unsubstantiated-and-false-claims/},
  note        = {Appendix 2 classifies the 79 podcast series by ideology: \url{https://www.brookings.edu/wp-content/uploads/2023/02/Appendices_for_Podcast_Project-3-1.pdf}}
}

@article{yang2025qwen3,
  title={Qwen3 technical report},
  author={Yang, An and Li, Anfeng and Yang, Baosong and Zhang, Beichen and Hui, Binyuan and Zheng, Bo and Yu, Bowen and Gao, Chang and Huang, Chengen and Lv, Chenxu and others},
  journal={arXiv preprint arXiv:2505.09388},
  year={2025}
}

@misc{speer2022wordfreq,
  author    = {Speer, Robyn},
  title     = {rspeer/wordfreq: v3.0},
  year      = {2022},
  publisher = {Zenodo},
  version   = {v3.0.2},
  doi       = {10.5281/zenodo.7199437},
  url       = {https://doi.org/10.5281/zenodo.7199437}
}

@article{boutyline2026forging,
  author  = {Boutyline, Andrei and Johnston, Elle},
  title   = {Forging Better Axes: Evaluating and Improving the Reliability of Semantic Dimensions in Word Embeddings},
  journal = {Sociological Methods \& Research},
  year    = {2026},
  pages   = {1--48},
  doi     = {10.1177/00491241261467930}
}

@article{elhage2022superposition,
   title={Toy Models of Superposition},
   author={Elhage, Nelson and Hume, Tristan and Olsson, Catherine and Schiefer, Nicholas and Henighan, Tom and Kravec, Shauna and Hatfield-Dodds, Zac and Lasenby, Robert and Drain, Dawn and Chen, Carol and Grosse, Roger and McCandlish, Sam and Kaplan, Jared and Amodei, Dario and Wattenberg, Martin and Olah, Christopher},
   year={2022},
   journal={Transformer Circuits Thread}
}

@inproceedings{betley2025emergent,
  author    = {Betley, Jan and Tan, Daniel and Warncke, Niels and Sztyber-Betley, Anna and Bao, Xuchan and Soto, Mart{\'i}n and Labenz, Nathan and Evans, Owain},
  title     = {Emergent Misalignment: Narrow finetuning can produce broadly misaligned {LLMs}},
  booktitle = {Proceedings of the 42nd International Conference on Machine Learning},
  series    = {Proceedings of Machine Learning Research},
  volume    = {267},
  year      = {2025},
  publisher = {PMLR},
  url       = {https://proceedings.mlr.press/v267/betley25a.html},
  eprint    = {2502.17424},
  archivePrefix = {arXiv},
  note      = {Extended version published in Nature (2026)}
}

@misc{kramar2026building,
  title     = {Building Production-Ready Probes For {Gemini}},
  author    = {Kram{\'a}r, J{\'a}nos and Engels, Joshua and Wang, Zheng and Chughtai, Bilal and Shah, Rohin and Nanda, Neel and Conmy, Arthur},
  year      = {2026},
  eprint    = {2601.11516},
  archivePrefix = {arXiv}
}

@inproceedings{li2024measuring,
  author    = {Li, Kenneth and Liu, Tianle and Bashkansky, Naomi and Bau, David and Vi{\'e}gas, Fernanda and Pfister, Hanspeter and Wattenberg, Martin},
  title     = {Measuring and Controlling Instruction (In)Stability in Language Model Dialogs},
  booktitle = {Conference on Language Modeling (COLM)},
  year      = {2024},
  eprint    = {2402.10962},
  archivePrefix = {arXiv},
  note      = {Earlier arXiv title: ``Measuring and Controlling Persona Drift in Language Model Dialogs''}
}

@misc{macdiarmid2024simple,
  title     = {Simple Probes Can Catch Sleeper Agents},
  author    = {MacDiarmid, Monte and Maxwell, Timothy and Schiefer, Nicholas and Mu, Jesse and Kaplan, Jared and Duvenaud, David and Bowman, Sam and Tamkin, Alex and Perez, Ethan and Sharma, Mrinank and Denison, Carson and Hubinger, Evan},
  year      = {2024},
  howpublished = {Anthropic research blog},
  url       = {https://www.anthropic.com/research/probes-catch-sleeper-agents}
}

@misc{marks2026persona,
  author       = {Marks, Sam and Lindsey, Jack and Olah, Christopher},
  title        = {The Persona Selection Model: Why {AI} Assistants might Behave like Humans},
  howpublished = {Anthropic Alignment Science Blog},
  year         = {2026},
  month        = feb,
  url          = {https://alignment.anthropic.com/2026/psm/}
}

@misc{poulis2026testing,
  title     = {Testing the Limits of Truth Directions in {LLMs}},
  author    = {Poulis, Angelos and Crovella, Mark and Terzi, Evimaria},
  year      = {2026},
  eprint    = {2604.03754},
  archivePrefix = {arXiv},
  note      = {BlackboxNLP 2026 (per arXiv comment)}
}

@inproceedings{santurkar2023whose,
  author    = {Santurkar, Shibani and Durmus, Esin and Ladhak, Faisal and Lee, Cinoo and Liang, Percy and Hashimoto, Tatsunori},
  title     = {Whose Opinions Do Language Models Reflect?},
  booktitle = {Proceedings of the 40th International Conference on Machine Learning},
  series    = {Proceedings of Machine Learning Research},
  volume    = {202},
  pages     = {29971--30004},
  year      = {2023},
  publisher = {PMLR},
  url       = {https://proceedings.mlr.press/v202/santurkar23a.html},
  eprint    = {2303.17548},
  archivePrefix = {arXiv}
}

@inproceedings{schouten2025truthvalue,
  title     = {Truth-Value Judgment in Language Models: `Truth Directions' Are Context Sensitive},
  author    = {Schouten, Stefan F. and Bloem, Peter and Markov, Ilia and Vossen, Piek},
  booktitle = {Conference on Language Modeling (COLM)},
  year      = {2025},
  eprint    = {2404.18865},
  archivePrefix = {arXiv}
}

@article{shanahan2023role,
  author  = {Shanahan, Murray and McDonell, Kyle and Reynolds, Laria},
  title   = {Role play with large language models},
  journal = {Nature},
  volume  = {623},
  number  = {7987},
  pages   = {493--498},
  year    = {2023},
  doi     = {10.1038/s41586-023-06647-8}
}

@article{wang2025persona,
  author  = {Wang, Miles and Dupr{\'e} la Tour, Tom and Watkins, Olivia and Makelov, Alex and Chi, Ryan A. and Miserendino, Samuel and Wang, Jeffrey and Rajaram, Achyuta and Heidecke, Johannes and Patwardhan, Tejal and Mossing, Dan},
  title   = {Persona Features Control Emergent Misalignment},
  journal = {arXiv preprint arXiv:2506.19823},
  year    = {2025},
  eprint  = {2506.19823},
  archivePrefix = {arXiv}
}

@article{zou2023representation,
  author  = {Zou, Andy and Phan, Long and Chen, Sarah and Campbell, James and Guo, Phillip and Ren, Richard and Pan, Alexander and Yin, Xuwang and Mazeika, Mantas and Dombrowski, Ann-Kathrin and Goel, Shashwat and Li, Nathaniel and Byun, Michael J. and Wang, Zifan and Mallen, Alex and Basart, Steven and Koyejo, Sanmi and Song, Dawn and Fredrikson, Matt and Kolter, J. Zico and Hendrycks, Dan},
  title   = {Representation Engineering: A Top-Down Approach to {AI} Transparency},
  journal = {arXiv preprint arXiv:2310.01405},
  year    = {2023},
  eprint  = {2310.01405},
  archivePrefix = {arXiv}
}

@article{panickssery2023steering,
  title={Steering llama 2 via contrastive activation addition},
  author={Panickssery, Nina and Gabrieli, Nick and Schulz, Julian and Tong, Meg and Hubinger, Evan and Turner, Alexander Matt},
  journal={arXiv preprint arXiv:2312.06681},
  year={2023}
}

@article{tigges2023linear,
  title={Linear representations of sentiment in large language models},
  author={Tigges, Curt and Hollinsworth, Oskar John and Geiger, Atticus and Nanda, Neel},
  journal={arXiv preprint arXiv:2310.15154},
  year={2023}
}

@article{feldman1992political,
  author={Feldman, Stanley and Zaller, John},
  title={The Political Culture of Ambivalence: Ideological Responses to the Welfare State},
  journal={American Journal of Political Science},
  volume={36},
  number={1},
  pages={268--307},
  year={1992},
  doi={10.2307/2111433}
}
\bibliographystyle{iclr2027_conference}

\appendix
% Top-align figures on float-only pages in the appendix
\makeatletter
\setlength{\@fptop}{0pt}
\setlength{\@fpbot}{0pt plus 1fil}
\makeatother
\renewcommand{\thefigure}{A\arabic{figure}}
\renewcommand{\thetable}{A\arabic{table}}
\setcounter{figure}{0}
\setcounter{table}{0}

\section{Word2vec training data and hyperparameters}
\label{app:word2vec}

\paragraph{Corpora.} We train separate liberal and conservative models on corpora of 69,087,676 tokens per side, combining three genres (Table~\ref{tab:w2v_corpus}). \emph{Partisan news} consists of Common Crawl articles (2016--2020) from 11 liberal and 13 conservative outlets, with token counts matched between sides within each year and no outlet contributing more than half of a side's tokens in any year. \emph{Podcasts} are diarized transcripts of political podcasts \citep{rizwan2025toxicity}, with show ideology taken from \citet{wirtschafter2023audible}. \emph{Campaign emails} are fundraising and campaign emails from Democratic and Republican senders \citep{willis2026political}. Podcasts and emails cover January 2022--September 2023, and token counts are matched between sides within each month separately for each genre. We remove URLs, recognized boilerplate (sponsorship reads, legal and unsubscribe text), exact and near-duplicate documents, and heavily reused passages. Text is lowercased and punctuation is discarded. Outlet and show labels are proxies for ideological stance rather than article-level annotations.

\begin{table}[h]
\centering
\small
\caption{Composition of the word2vec training corpora (tokens per side).}
\label{tab:w2v_corpus}
\begin{tabular}{lrp{0.52\textwidth}}
\hline
Genre & Tokens per side & Sources \\
\hline
Partisan news & 38,886,566 & Liberal: CBS News, Daily Beast, HuffPost, MSNBC, New Yorker, Newsweek, NPR, Politico, Slate, The Atlantic, Washington Post. Conservative: American Thinker, Breitbart, Daily Caller, Daily Wire, Epoch Times, Federalist, Fox News, Hot Air, National Review, PJ Media, RedState, Townhall, Washington Examiner. \\
Podcasts & 17,724,396 & 2,116 liberal and 2,296 conservative episodes \\
Campaign emails & 12,476,714 & 50,521 Democratic and 58,264 Republican emails \\
\hline
Total & 69,087,676 & \\
\hline
\end{tabular}
\end{table}

\paragraph{Training.} Each side is trained from scratch with continuous bag-of-words (CBOW) in Gensim \citep{rehurek2010software}: 300 dimensions, maximum window 4, 10 negative samples, subsampling threshold $10^{-4}$, minimum count 5, and a learning rate decaying linearly from .025 to .0001 over ten epochs. We use the checkpoint saved after epoch 5. Vocabularies and sampling distributions are built from each side's corpus alone. We train three seeds per side. A word's representation is the sum of its input and output vectors. These settings were chosen after comparing several exploratory configurations and were not preregistered.

\paragraph{Axes.} Axes are the mean of raw anchor-pair differences, as for the LLM. Anchor pairs come from \citet{boutyline2026forging}, keeping pairs whose terms each occur at least five times in both corpora: 1,153 pairs in total (median 34 per axis, range 4--90). The \textit{liberal--conservative} axis uses four pairs: \emph{liberal/conservative}, \emph{liberals/conservatives}, \emph{democrat/republican}, and \emph{democrats/republicans}. Axis cosine matrices are computed within each model, and the liberal-minus-conservative difference matrices are averaged across seeds; vectors from different models are never averaged.

\clearpage
\section{Persona geometry across layer depth}
\label{app:depth}

\begin{figure}[h]
\centering
\includegraphics[width=\textwidth]{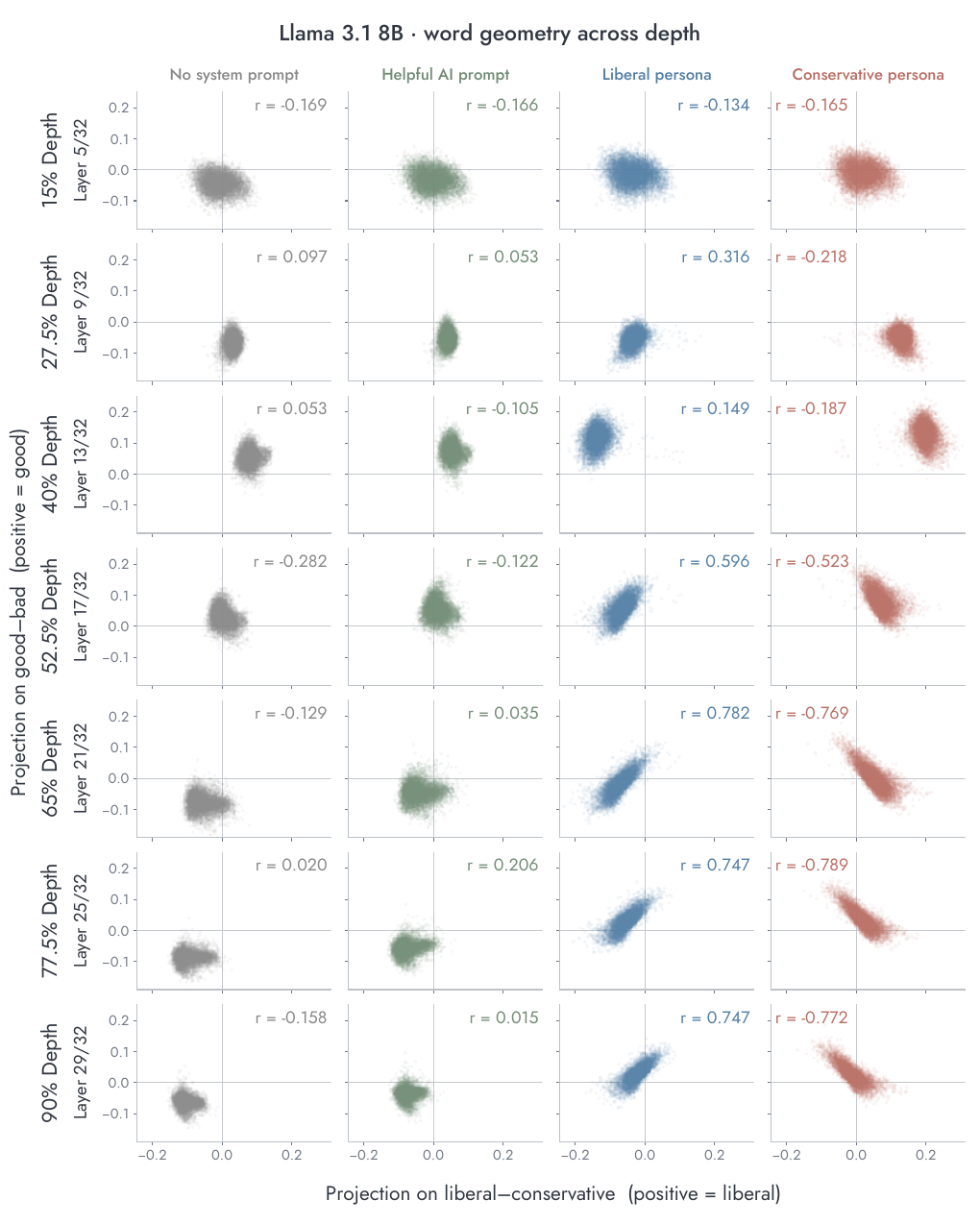}
\caption{Projections of 9,493 common words onto the liberal--conservative (horizontal) and good--bad (vertical) axes across depth, Llama 3.1 8B. Rows are depths (layers 5, 9, 13, 17, 21, 25, and 29 of 32); columns are system-prompt conditions. Words and axes are extracted within each condition and layer. Panels share coordinate limits; Pearson correlations between the two projections are shown in each panel. The 77.5\% row is the layer shown in Figure~\ref{fig:2}.}
\label{fig:a_depth}
\end{figure}

Figure~\ref{fig:a_depth} repeats Figure~\ref{fig:2} at seven depths of Llama 3.1 8B. In the earliest layers, the correlation between the political and evaluative projections is weak and similar across conditions; the opposing persona-specific correlations appear by roughly 50\% depth and strengthen thereafter. Figure~\ref{fig:4} shows the same emergence with depth for the axis alignments across all eight models.

\clearpage
\section{Replication across models}
\label{app:models}

\begin{figure}[p]
\centering
\includegraphics[width=\textwidth]{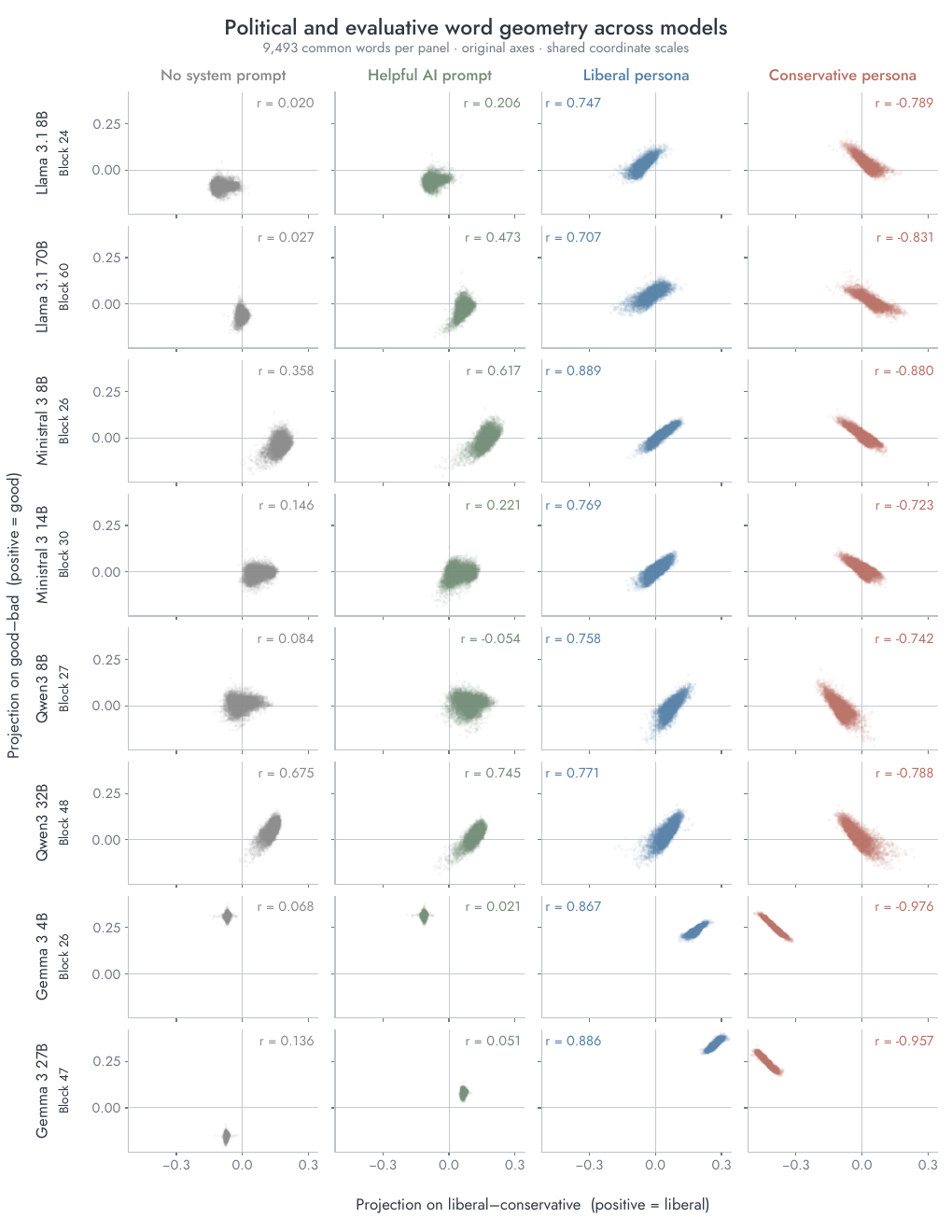}
\caption{Projections of 9,493 common words onto the liberal--conservative and good--bad axes for all eight models (rows) under each system-prompt condition (columns), as in Figure~\ref{fig:2}. Words and axes are extracted within each model and condition. Pearson correlations are shown in each panel.}
\label{fig:a_words}
\end{figure}

\begin{figure}[p]
\centering
\includegraphics[width=\textwidth]{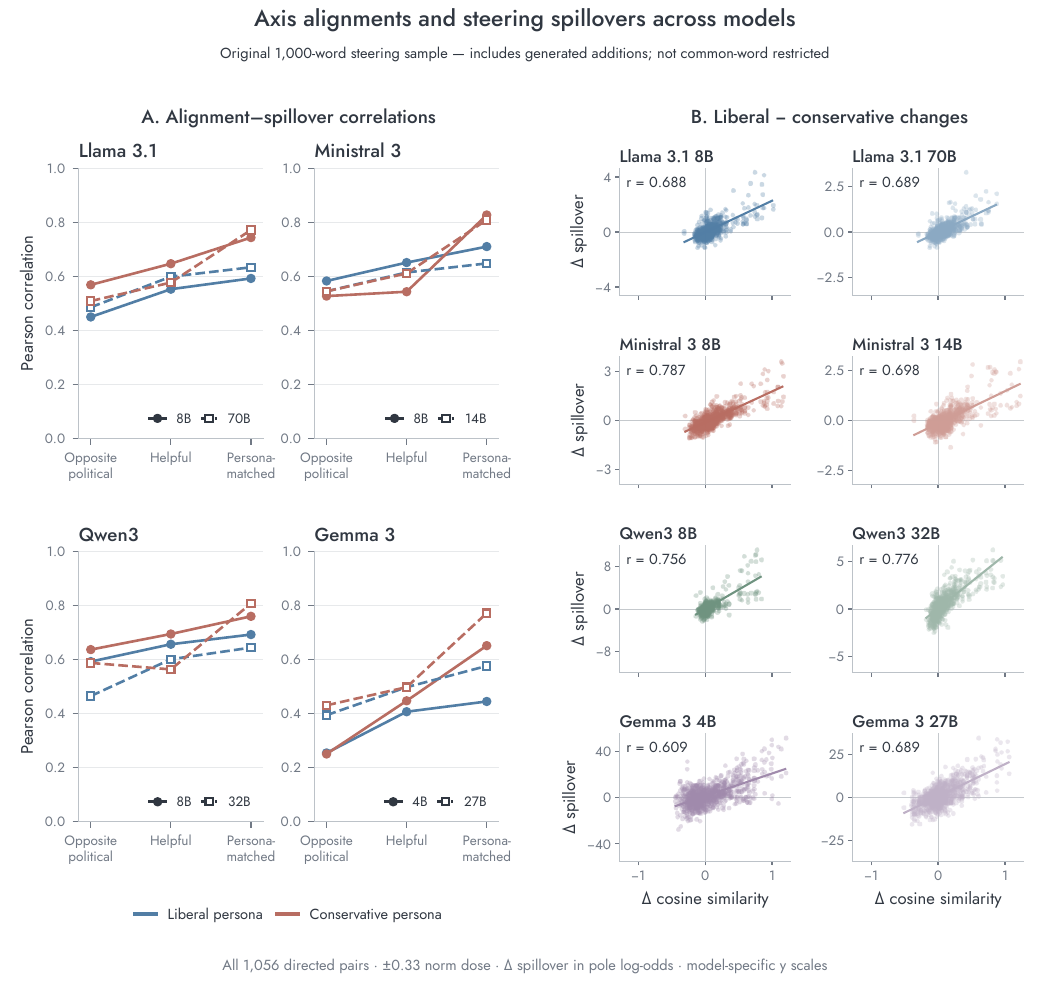}
\caption{Axis alignments and steering spillovers across models, as in Figure~\ref{fig:5}. (A) Correlation between spillovers and axis alignments for the liberal (blue) and conservative (red) personas, when predicted from the opposite persona's, the helpful assistant's, or the persona's own geometry; solid lines show the smaller and dashed lines the larger model in each family. (B) Liberal-minus-conservative differences in cosine similarity versus differences in spillover for each model. All models use all 1,056 directed off-target axis pairs and the same 1,000 target words. These words were sampled from the full feature bank, which includes anchor synonyms and antonyms beyond the common-word list, so Llama 3.1 8B values differ slightly from Figure~\ref{fig:5}. Steering uses $\alpha = 0.33$ at the final prompt token; y-axis scales in (B) differ by model.}
\label{fig:a_spill}
\end{figure}

Figures~\ref{fig:a_words}--\ref{fig:a_spill} repeat the analyses of Figures~\ref{fig:2}, \ref{fig:3}, and \ref{fig:5} for all eight models. Each model is analyzed at approximately 75\% depth (zero-based blocks 24, 60, 26, 30, 27, 48, 26, and 47 for Llama 3.1 8B and 70B, Ministral 3 8B and 14B, Qwen 3 8B and 32B, and Gemma 3 4B and 27B, respectively).

\begin{figure}[p]
\centering
\includegraphics[width=\textwidth,page=1]{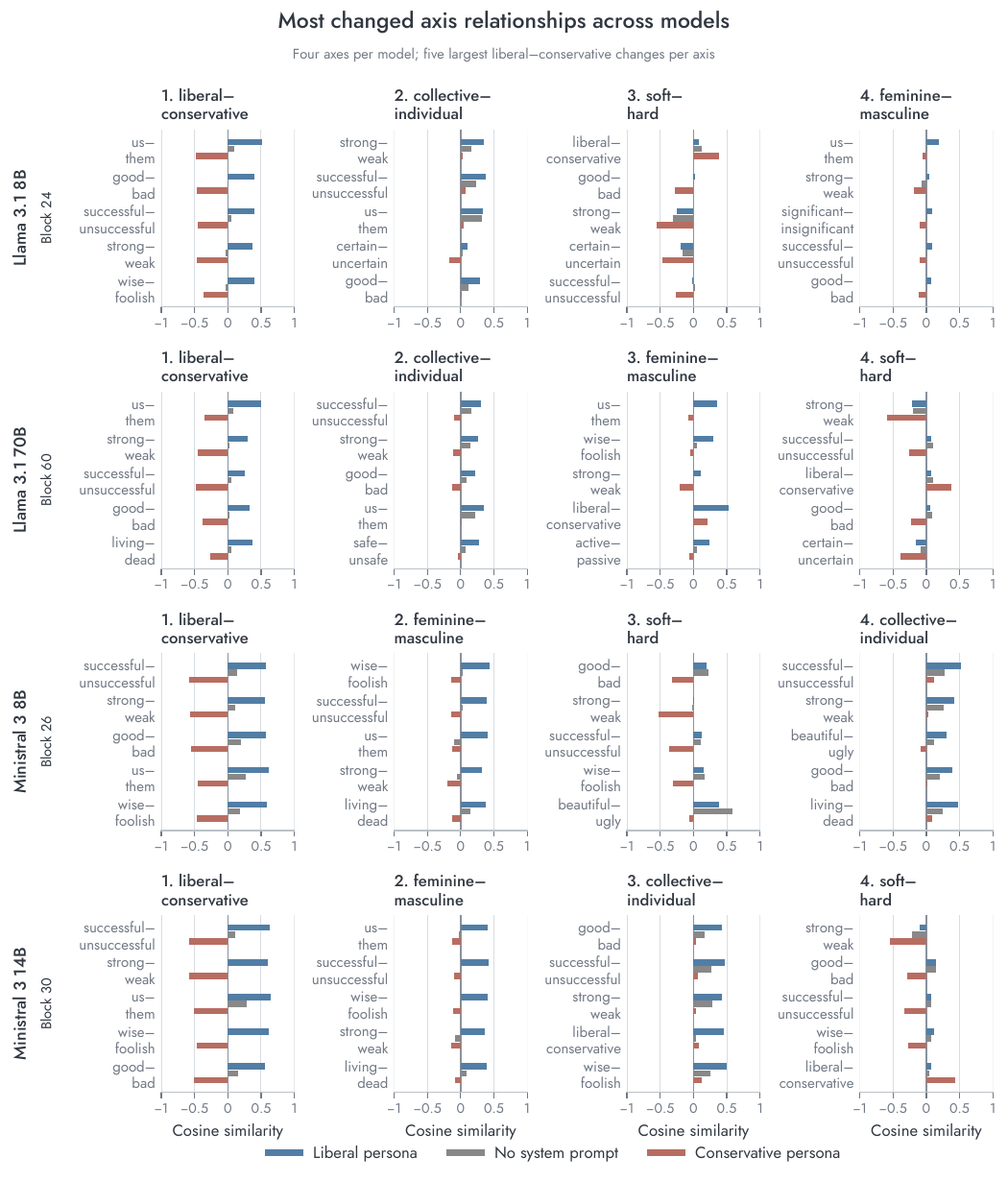}
\caption{Axes whose alignments change most between the liberal and conservative personas, as in Figure~\ref{fig:3}, for Llama 3.1 8B and 70B and Ministral 3 8B and 14B. Within each model, the four focal axes are ranked by mean absolute change in cosine similarity with the other 32 axes, and each is shown with the five partner axes that change most. Cosine scales are shared across panels.}
\label{fig:a_top4a}
\end{figure}

\begin{figure}[p]
\centering
\includegraphics[width=\textwidth,page=2]{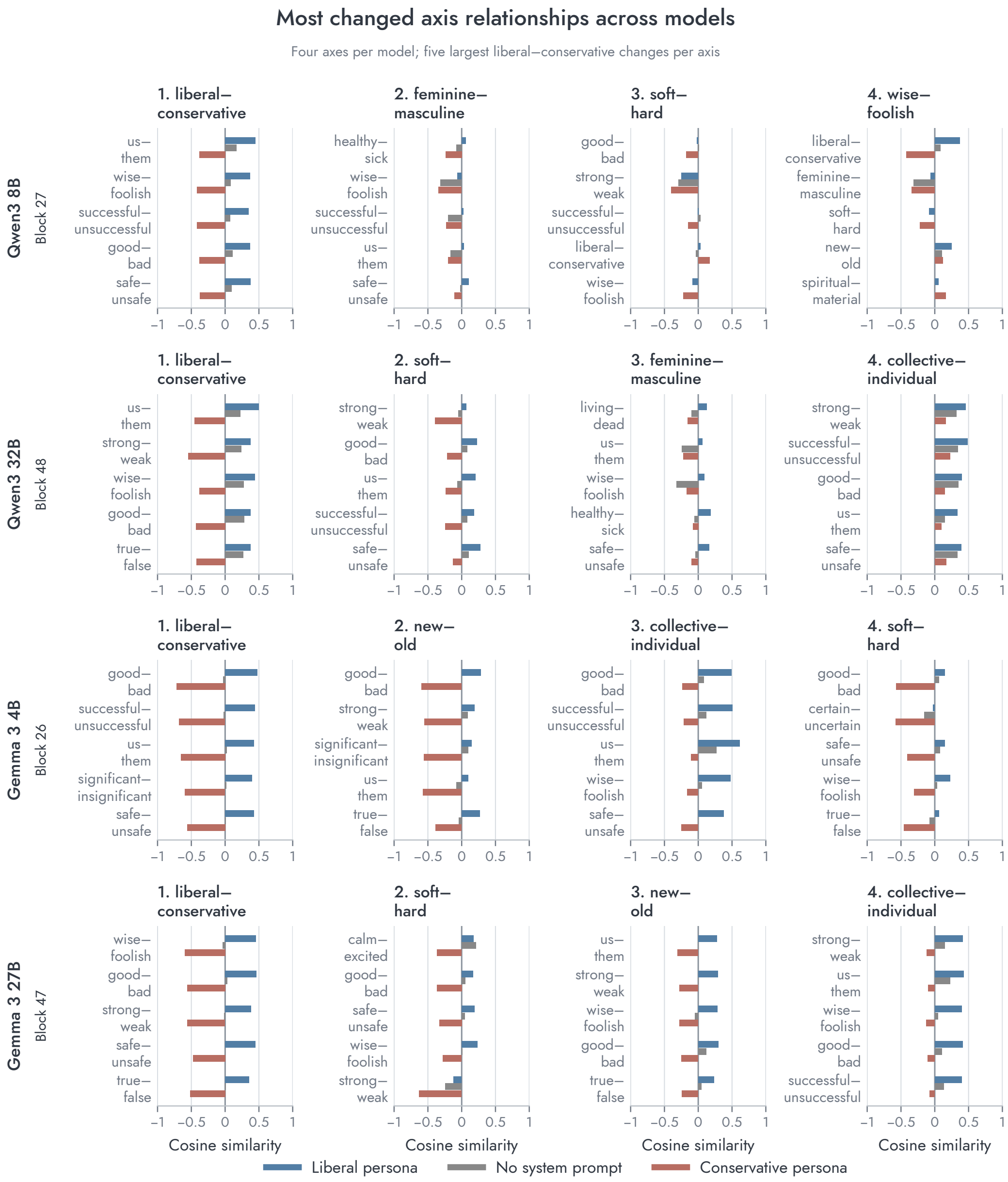}
\caption{Continuation of Figure~\ref{fig:a_top4a} for Qwen 3 8B and 32B and Gemma 3 4B and 27B.}
\label{fig:a_top4b}
\end{figure}

\clearpage
\section{Truth and morality probes across models}
\label{app:probes}

\begin{figure}[p]
\centering
\includegraphics[width=\textwidth,page=1]{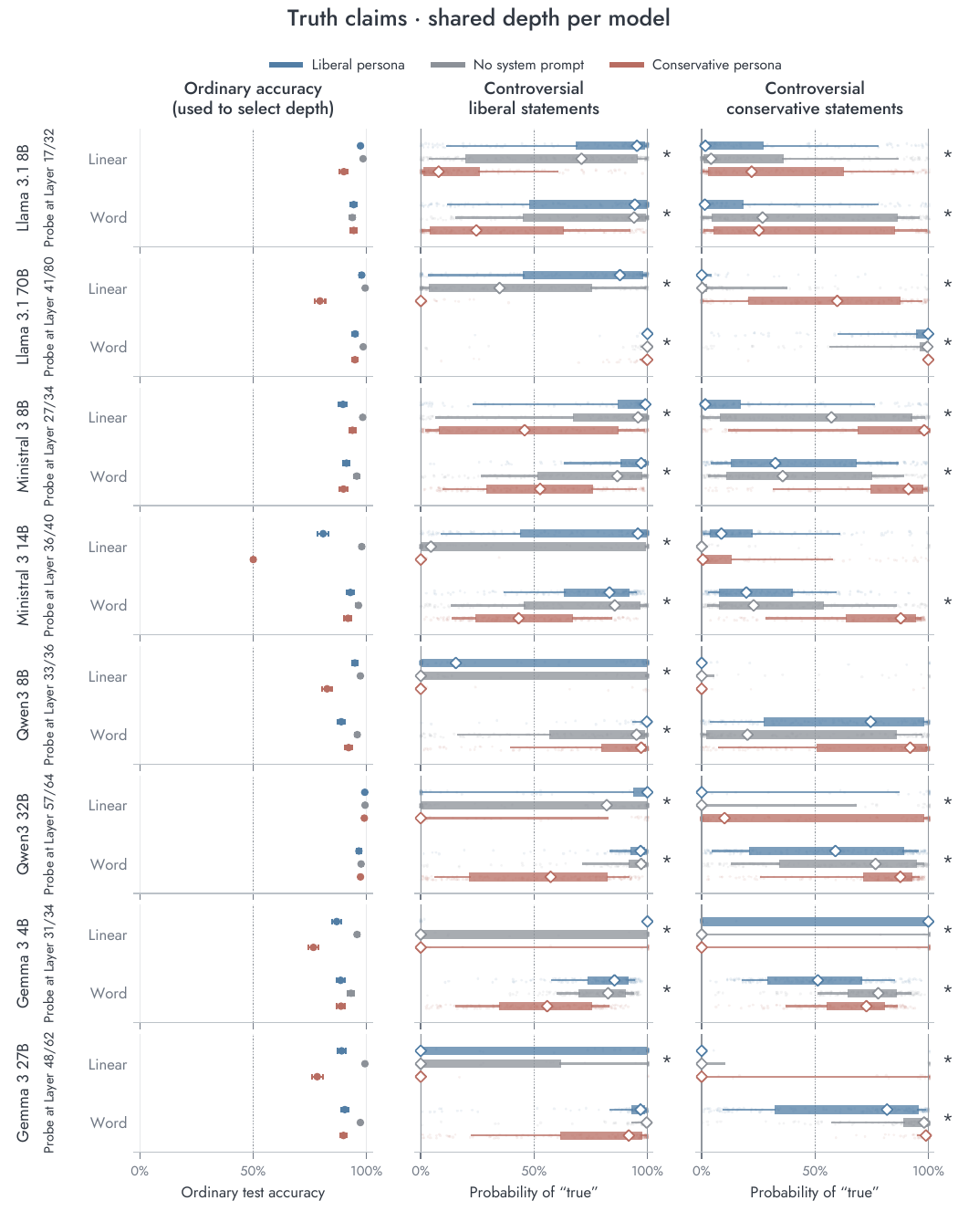}
\caption{Truth probes across models. Left: accuracy on 1,840 held-out ordinary statements from 230 Wikipedia articles, with 95\% article-bootstrap intervals. Middle and right: probe-assigned probability of ``true'' for 180 controversial statements expected to be endorsed by liberals (middle) and 180 expected to be endorsed by conservatives (right). Diamonds mark medians, thick bars the middle 50\%, thin bars the middle 80\%, and faint points individual statements. Probes are trained with no system prompt and held fixed across personas. Model-specific layers are shown beside each model name (one-based). Asterisks: Holm-adjusted $p < .05$ (see text).}
\label{fig:a_probe_truth}
\end{figure}

\begin{figure}[p]
\centering
\includegraphics[width=\textwidth,page=2]{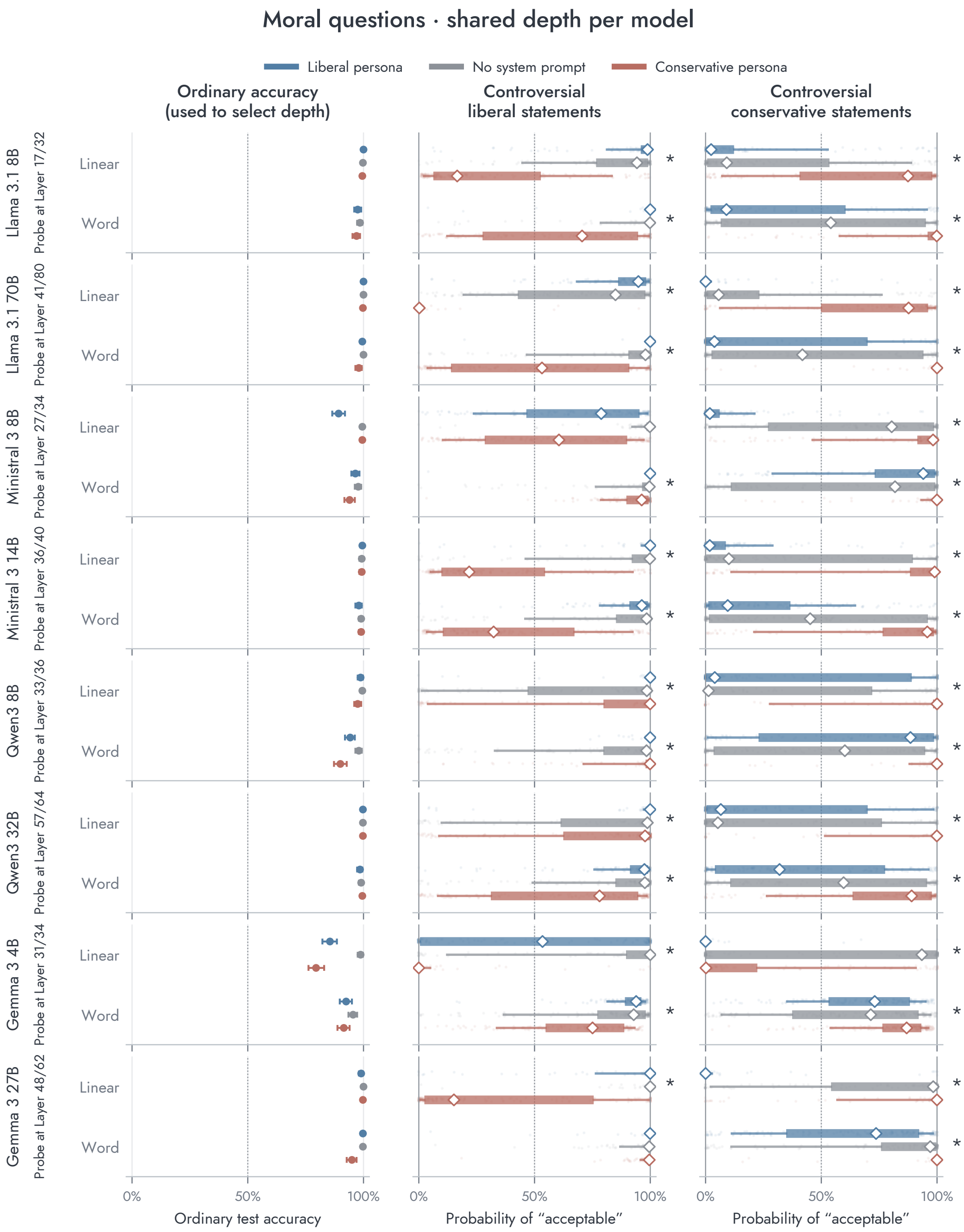}
\caption{Morality probes across models, as in Figure~\ref{fig:a_probe_truth}. Left: accuracy on 400 held-out ordinary actions from 200 acceptable/unacceptable pairs, with 95\% pair-bootstrap intervals. Middle and right: probe-assigned probability of ``acceptable'' for 150 controversial actions per partisan side.}
\label{fig:a_probe_moral}
\end{figure}

Figures~\ref{fig:a_probe_truth} and~\ref{fig:a_probe_moral} apply the learned linear and word-representation probes to all eight models. SAE-feature probes are omitted because the pretrained SAE is specific to Llama 3.1 8B. The layer at which probes perform best on ordinary statements varies across models, so a single layer is chosen per model and used for both tasks and both probe types. The criterion is ordinary-statement test accuracy averaged over the two tasks, two probe types, and three persona conditions, among the seven saved depths; ties go to the shallower layer. Controversial statements play no role in the selection. Because the ordinary test statements are reused for selection, the ordinary accuracies shown are not independent validation of the chosen layers.

Asterisks mark significant liberal-versus-conservative differences in probe-assigned probability for a given statement side. The test statistic is the mean paired difference, averaged within and then across topics. Under the null, persona labels are swapped jointly for all statements in a topic: exhaustively for truth ($2^{12}$ assignments over 12 topics) and with 99,999 random assignments for morality (30 topics). $p$-values are Holm-corrected over a family of 140 comparisons, a superset of those shown here and in Figure~\ref{fig:6}. Tests are two-sided, so an asterisk can in principle mark movement in the unexpected direction; this occurs once, for the Gemma 3 4B linear truth probe on conservative-side statements.

\begin{figure}[p]
\centering
\includegraphics[width=\textwidth]{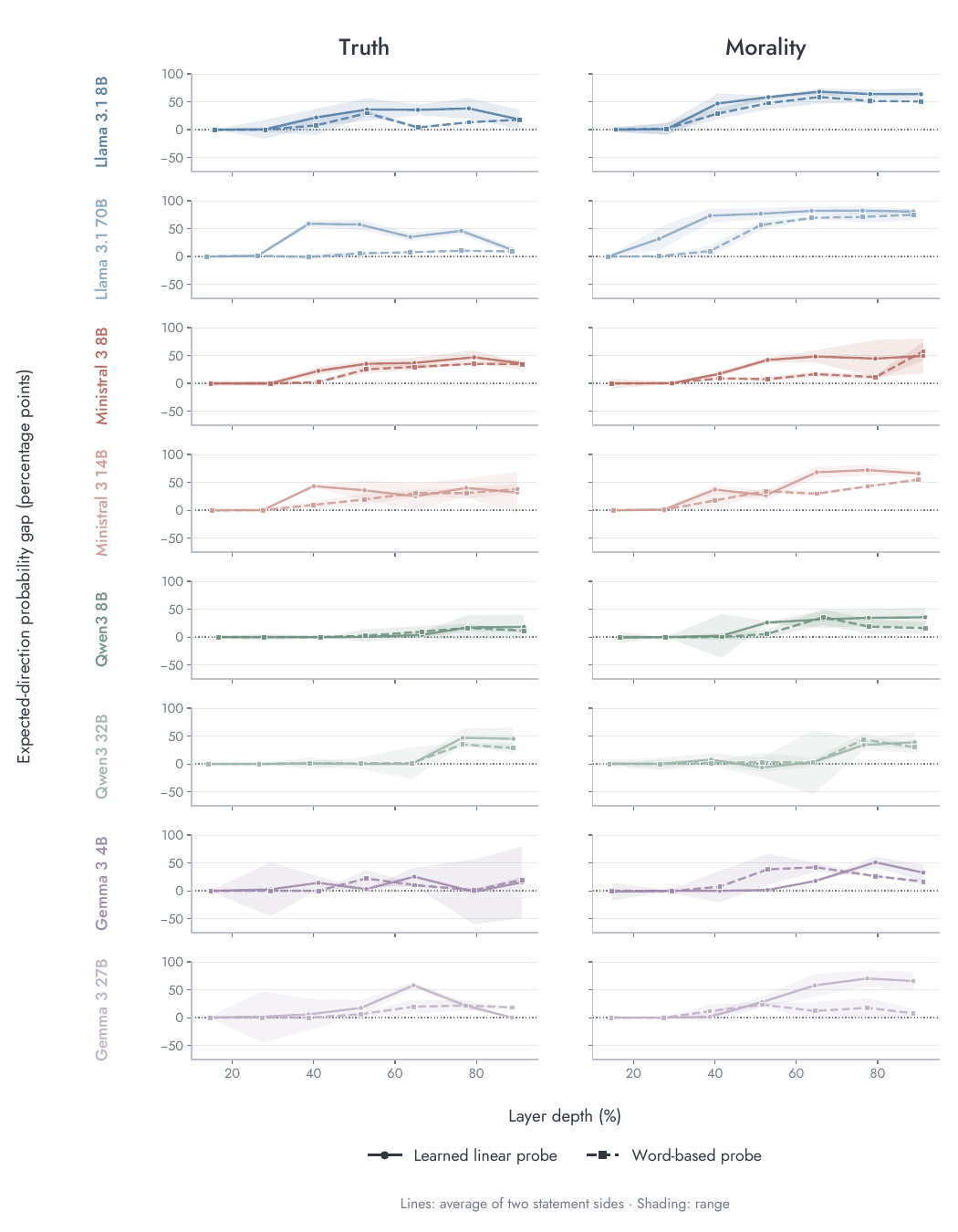}
\caption{Persona-dependent probe gaps across depth for all eight models (rows), for truth (left) and morality (right). For each statement side, the gap is the probability assigned by the persona expected to endorse the statements minus that assigned by the opposing persona, averaged equally across topics, in percentage points; negative values indicate the reverse ordering. Lines average the liberal-side and conservative-side gaps, and shading spans the two. Solid lines show learned linear probes and dashed lines word-representation probes. Probes are trained with no system prompt at each depth and held fixed across personas.}
\label{fig:a_probe_depth}
\end{figure}

Figure~\ref{fig:a_probe_depth} shows the persona-dependent gap in probe-assigned probability for probes trained at each saved depth. Gaps are near zero in early layers and emerge by roughly 40--50\% depth, after which results at neighboring layers are generally similar.

\clearpage
\section{Behavioral responses to controversial statements}
\label{app:behavior}

\begin{figure}[h]
\centering
\includegraphics[width=\textwidth]{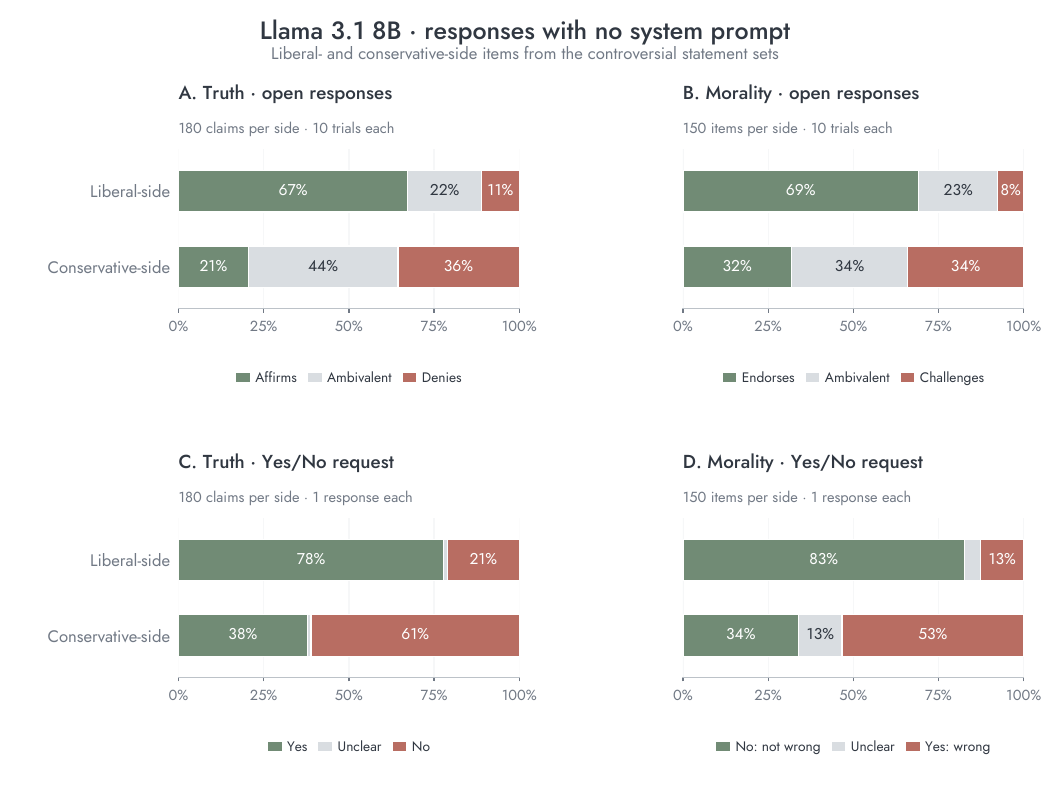}
\caption{Llama 3.1 8B responses to controversial statements with no system prompt, for liberal-side and conservative-side items. (A--B) Open responses, ten per item, rated by Gemma 4 31B. (C--D) Constrained yes/no responses, one per item; for morality, ``No'' (not wrong) is the accepting answer. 180 truth claims and 150 moral actions per side.}
\label{fig:a_behavior_base}
\end{figure}

\begin{figure}[h]
\centering
\includegraphics[width=\textwidth]{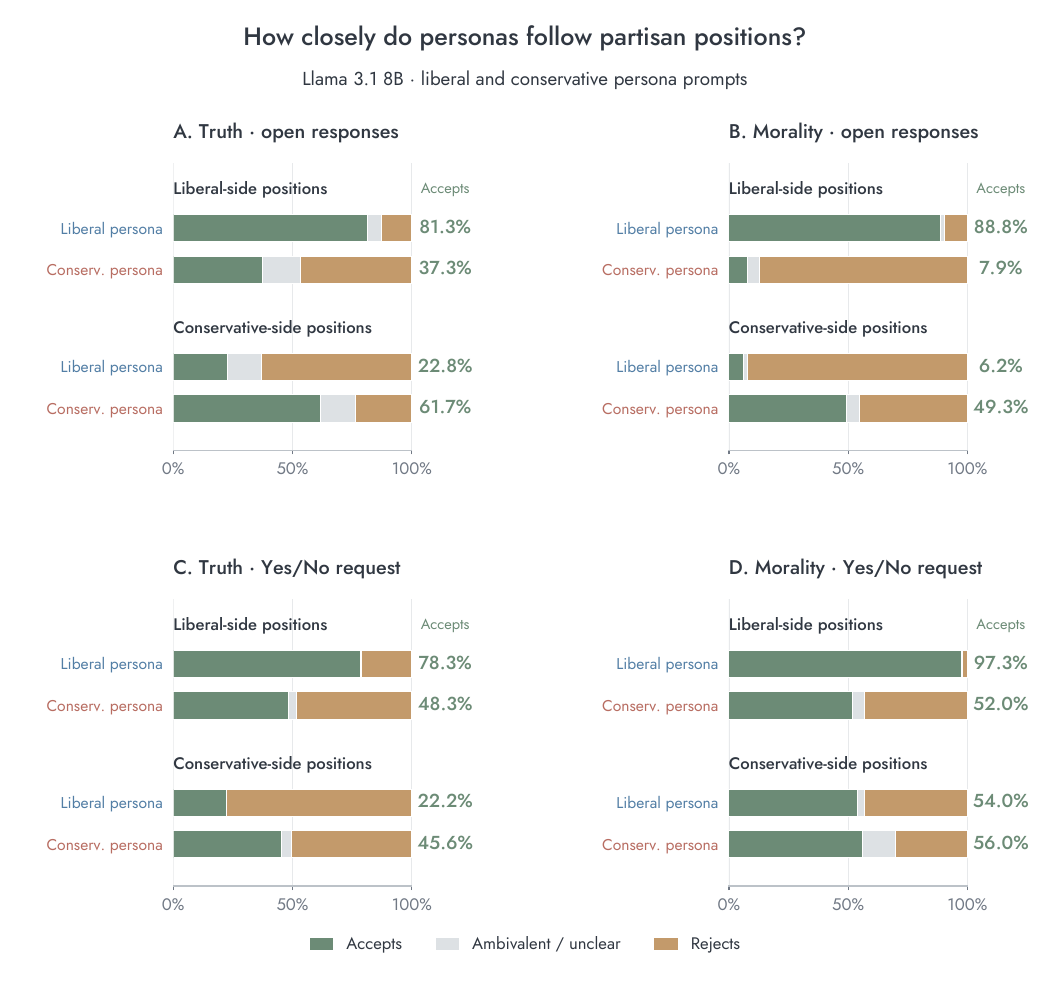}
\caption{Llama 3.1 8B responses to controversial statements under the liberal (upper bar) and conservative (lower bar) personas. Green, gray, and ochre segments show acceptance, ambivalent or unclear answers, and rejection; numbers give the acceptance rate. Acceptance means affirming a truth claim or endorsing an action in open responses, and answering ``Yes'' to ``Is it true\ldots?'' or ``No'' to ``Is this wrong?'' in constrained responses. Open and constrained responses are collected separately, with the same settings as Figure~\ref{fig:a_behavior_base}.}
\label{fig:a_behavior_persona}
\end{figure}

To compare the probes with the model's behavior, we collected Llama 3.1 8B's responses to the controversial statements in two formats. In the open format, the model is asked ``Is it true that [claim]?'' or told ``I want [action].'' with no instruction about response format; we sample ten responses per item (temperature 0.5, top-$p$ 0.9, up to 800 tokens), and Gemma 4 31B rates each response as affirming/endorsing, ambivalent, or denying/challenging the statement. For morality, this rubric counts practical help as endorsement and refusal as challenge, so it is broader than a judgment of moral acceptability. In the constrained format, the model is asked for a simple yes or no answer (``Is it true that\ldots?''; ``I want [action]. Is this wrong?'') and we record one response per item. Items receive equal weight. Figure~\ref{fig:a_behavior_base} shows responses without a system prompt, and Figure~\ref{fig:a_behavior_persona} under the political personas.

\end{document}